\documentclass[preprint,8pt,authoryear]{elsarticle}

\usepackage{amssymb}
\usepackage{lipsum}

\usepackage{fullpage}
\usepackage{wrapfig}
\usepackage{comment}
\usepackage[utf8]{inputenc} 

\usepackage[T1]{fontenc}    
\usepackage{hyperref}       
\usepackage{url}            
\usepackage{booktabs}       
\usepackage{amsfonts}       
\usepackage{nicefrac}       
\usepackage{microtype}      
\usepackage{xcolor}         
\usepackage{booktabs} 
\usepackage{siunitx} 
\usepackage{graphicx} 
\usepackage{xcolor} 
\definecolor{bestcolor}{rgb}{1,0,0} 
\usepackage{subcaption}
\usepackage{caption}
\usepackage[normalem]{ulem}
\usepackage{tikz}
\usepackage{comment}
\usepackage{amsfonts}       
\usepackage{amssymb}
\usepackage{amsmath}
\usepackage{amsthm}
\usepackage{mathtools}
\usepackage{multicol}
\usepackage{multirow}

\usepackage{color}
\usepackage[ruled,vlined]{algorithm2e}

\usepackage{array}
\usepackage[shortlabels]{enumitem}

\usepackage{longtable}
\usepackage{geometry}
\usepackage{subcaption}
\usepackage{hyperref}
\graphicspath{{./Figures/} } 
\usepackage[normalem]{ulem}
\newcounter{ToDo}
\newcounter{gaocomm} 
\newcounter{Note}
\definecolor{blue-violet}{rgb}{0.00,0.75,0.90}
\definecolor{mygreen}{rgb}{0.0, 0.5, 0.0}
\definecolor{awesome}{rgb}{1.0, 0.13, 0.32}
\definecolor{bostonuniversityred}{rgb}{1.0, 0.0, 0.0}

\usepackage{hyperref}
\usepackage[symbol]{footmisc}

\theoremstyle{definition}

\journal{Not sure.}

\begin{document}

\begin{frontmatter}



\title{A Spatial-Temporal Large Language Model with Diffusion (STLLM-DF) for Enhanced Multi-Mode Traffic System Forecasting}

\author[first]{Zhiqi Shao\fnref{equalcontrib}}
\ead{zhiqi.shao@sydney.edu.au}
\author[second]{Haoning Xi\fnref{equalcontrib}}
\ead{alice.xi@newcastle.edu.au}
\author[third]{Haohui Lu}
\ead{haohui.lu@sydney.edu.au}
\author[first]{Ze Wang\corref{cor1}}
\ead{ze.wang@sydney.edu.au}
\author[first]{Michael G.H. Bell}
\ead{michael.bell@sydney.edu.au}
\author[first]{Junbin Gao}
\ead{junbin.gao@sydney.edu.au}

\cortext[cor1]{Corresponding author}
\fntext[equalcontrib]{Zhiqi Shao and Haoning Xi contributed equally to this work.}

\affiliation[first]{organization={The University of Sydney},
            addressline={Business Analytics Discipline}, 
            city={Camperdown},
            postcode={NSW 2006}, 
            state={New South Wales},
            country={Australia}}

\affiliation[second]{organization={The University of Newcastle},
            addressline={Newcastle Business School}, 
            city={Newcastle},
            postcode={NSW 2300}, 
            state={New South Wales},
            country={Australia}}

\affiliation[third]{organization={The University of Sydney},
            addressline={Faculty of Engineering}, 
            city={Camperdown},
            postcode={NSW 2006}, 
            state={New South Wales},
            country={Australia}}

\begin{abstract}
The rapid advancement of Intelligent Transportation Systems (ITS) presents challenges, particularly with missing data in multi-modal transportation data and the complexity of handling diverse sequential tasks within a centralized framework. To address these issues, we propose the Spatial-Temporal Large Language Model Diffusion (STLLM-DF), an innovative model that leverages Denoising Diffusion Probabilistic Models (DDPMs) and Large Language Models (LLMs) to improve multi-task transportation prediction. The DDPM's strong denoising capabilities allow it to recover underlying data patterns from noisy inputs, making it particularly effective in complex transportation systems. Meanwhile, the non-pretrained LLM adapts dynamically to spatial-temporal relationships within multi-modal networks, allowing the system to efficiently manage diverse transportation tasks in both long-term and short-term predictions. Extensive experiments demonstrate that STLLM-DF consistently outperforms existing models, achieving an average reduction of 2.40\% in MAE, 4.50\% in RMSE, and 1.51\% in MAPE. This model represents a significant advancement in centralized ITS, enhancing predictive accuracy, robustness, and overall system performance across multiple tasks.
Enhancing Spatio-Temporal Traffic Forecasting with Frozen Transformer Language Models and Diffusion Techniques 
\end{abstract}




\end{frontmatter}




\section{Introduction}
\label{introduction}

Intelligent Transportation Systems (ITS) are pivotal in enhancing modern transportation networks' efficiency, safety, and sustainability, overseeing critical functions like incident management, traffic flow optimization, and demand-responsive transport service management. However, these sophisticated systems are increasingly challenged by 
\begin{itemize}
    \item \textbf{Missing Values in Data:} The vast datasets employed by ITS often contain inherent missing values, or what we call noise. This noise can stem from sensor malfunctions, incomplete data transmission, or even environmental factors like weather. The presence of such noise severely hampers the generalization capabilities of traditional models, leading to inaccurate traffic predictions and suboptimal decision-making. Addressing data quality is, therefore, fundamental to improving predictive accuracy and the robustness of ITS \cite{WANG2019144TRC, dataquality}.
    \item \textbf{Sequential Multi-Task Data Handling:}  ITS relies on various data streams from multiple modes of transportation, including buses, taxis, and metro systems. Each of these data sources has its own unique temporal and spatial characteristics, creating a highly complex dataset that needs to be processed simultaneously. Integrating and managing these diverse datasets is challenging, particularly when missing value is present. Existing models often struggle to maintain accuracy and reliability across different tasks when handling such heterogeneous data streams \cite{dataquality2, Zannat2019dataquality}.
\end{itemize}
The above challenges highlight the need for a more robust approach to processing and utilizing transportation data. A model that can centralize the management of these diverse tasks while employing advanced noise reduction (missing value handling) techniques is critical to preserving data integrity and maximizing utility, ultimately improving the system's overall effectiveness and reliability. As shown in Figure~\ref{fig:intro} such a model would need to be capable of handling the complexity and scale of ITS tasks in real-time environments.

To address these challenges, this paper introduces the \textbf{Spatial-Temporal Large Language Model Diffusion (STLLM-DF)}, a novel approach that combines Denoising Diffusion Probabilistic Models (DDPMs) and Large Language Models (LLMs). DDPMs are generative models based on stochastic processes, simulating the diffusion of data from order to disorder and then reversing the process to restore the original data, effectively achieving advanced data recovery. LLMs are leveraged for high-level feature extraction, enabling the STLLM-DF model to eliminate the need for manual feature engineering or prompting mechanisms. This streamlined approach adapts effortlessly to large and complex datasets, significantly improving the accuracy of Intelligent Transportation System (ITS) operations.

This innovation marks the first application of such a model in ITS, designed to tackle multiple tasks concurrently. By utilizing advanced data recovery techniques, the STLLM-DF effectively mitigates the impact of noise on data quality, ensuring that accurate and reliable traffic information is delivered to support real-time decision-making. Additionally, using LLMs for feature extraction allows for a better understanding of transportation networks' spatial and temporal dynamics, enabling the model to respond to a wider array of transportation demands.
By incorporating these advanced data processing techniques and a deep understanding of ITS operational demands, our method offers a revolutionary solution that promises to significantly enhance system performance.

\subsection{Related Works}
In the recent studies have extensively applied machine learning and deep learning to predict various tasks to assist the intelligent transportation systems, including traffic flow \cite{shao2024ccdsreformertrafficflowprediction, shao2024stmambasynccomplementmambatransformers, shao2024stmambaspatialtemporalselectivestate}, traffic speed \cite{ZOU2023104263TRC, REMPE2022103448TRC}, taxi demand \cite{KIM2020102786TRC}, and accidents, as well as city bike traffic \cite{LI2023103984TRC}. Despite these advancements, developing models that perform uniformly well across these diverse tasks remains limited. This is primarily due to challenges associated with handling the variety of data types inherent in comprehensive traffic system analyses. The study \citep{6930534} was the first one to propose a method of unified model to make traffic analysis. It presents a traffic analysis platform that integrates a conceptual model with symbolic contexts to analyze traffic dynamics. 
 \cite{8525272} introduces a unified spatial-temporal model for short-term road traffic prediction that outperforms traditional methods by integrating physically intuitive but only applies to traffic flow tasks; thus, finding a unified model that can adopt multi-tasks is crucial. 

\subsubsection{ Denoising Diffusion Probabilistic Models (DDPMs) }
The diffusion model employed in this study is a generative model based on stochastic processes. It simulates the diffusion process of data from order to disorder and then reverses this process to restore the original data, thereby achieving data recovery. Compared to traditional denoising or data recovery techniques, the diffusion model has the advantage of powerful data recovery capabilities, especially when dealing with data with complex dependencies. The DDPMs have been widely used in generative tasks in different fields such as computer vision, medicine, etc. Recently, DDPMs have started to be used in traffic fields such as traffic data sampling \cite{10575606}, traffic anomaly detection \cite{LI2024111387}, and traffic matrix estimation \cite{10218016}. However, none of them are used to conduct a unified framework for adopting multi-task prediction in ITS.  

Our proposed STLLM-DF, which combines spatial temporal features, offers a new theoretical perspective and practical approach to processing traffic data.
The training objective is to minimize the difference between the noise predicted by the model and the actual noise added. In this way, the model learns how to accurately recover the original data from the high missing ratio data, i.e., perform data recovery. This training method enables the model to generate or restore high-quality data, particularly when given inputs that are partially damaged or interfered with by noise.

\subsubsection{Large Language Model (LLM)}

Large Language Models (LLMs) are increasingly utilized in the field of traffic management for diverse tasks such as multimodal traffic accident forecasting \cite{s23229225}, traffic prediction \cite{liu2024spatialtemporallargelanguagemodel}, and signal control \cite{pang2024illmtscintegrationreinforcementlearning}. Traditionally, these applications rely on specific prompts as inputs to guide the model's performance, which can complicate the framework and increase computational demands. Given these complexities, this paper critically examines whether LLMs can be effectively employed without prompt-based inputs while still maintaining robust performance. This investigation is pivotal for streamlining traffic management applications, potentially reducing computational overhead and simplifying the operational framework.

\subsection{Contribution}
We have identified a gap in existing models' ability to manage multiple tasks within a unified framework due to varying data quality. To address this, we propose the STLLM-DF framework, which incorporates Denoising Diffusion Probabilistic Models (DDPMs) for data denosing, followed by the use of a Large Language Model (LLM) to extract key traffic information. To the best of our knowledge, this innovative approach uniquely combines these technologies to enhance the processing and utility of traffic data across different tasks.
\begin{enumerate}
\item  \textbf{We are the pioneers in merging the capabilities of DDPM  with large language models to establish a groundbreaking model for a centralized ITS.} This innovative combination leverages the strengths of both technologies to tackle complex traffic management challenges.

\item \textbf{The DDPM is effective for multi-task transportation prediction due to its strong denoising capabilities and ability to recover underlying data patterns.} Our research takes a novel approach by utilizing DDPMs with time-step embedding named ST-Denoisng block, which not only can capture the dynamic changes of traffic data over time but also can do data recovery process, while the use of convolutional networks effectively captures the spatial characteristics of the data. This application significantly enhances data quality across various ITS tasks, making it a first in the field.

\item \textbf{For a centralized transportation system, the non-pretrained LLM demonstrates adaptability to varying spatial-temporal relationships.} Despite the lack of pretraining, the model can dynamically capture and learn from the changing spatial and temporal patterns present in multi-modal transportation networks, making it highly versatile in these environments.

\item \textbf{Extensive experiments across multiple tasks have demonstrated that STLLM-DF consistently outperforms other models.} By effectively leveraging the strengths of diffusion mechanisms and LLMs, STLLM-DF excels in handling the complexities of multi-task transportation forecasting, leading to superior accuracy and robustness compared to other approaches.

\end{enumerate}

The organization of this paper is as follows: Section~\ref{Preliminary} introduces the preliminaries, providing foundational concepts and essential background on Denoising diffusion probabilistic models and the Large Language model used in this study. Section~\ref{Method} presents the model architecture and methodology, detailing the design of the STLLM-DF model. Section~\ref{Experiment} outlines the experimental setup, including datasets, evaluation metrics, and implementation details; the vast model performance analysis is also included in this section. Section~\ref{Discussion and Implementation} provides an insight into the results, highlighting the model's performance and contributions to centralized ITS. Finally, the conclusion is presented in Section~\ref{conclusion}.

\begin{figure}[htbp!]
    \centering
    \includegraphics[scale = 0.3]{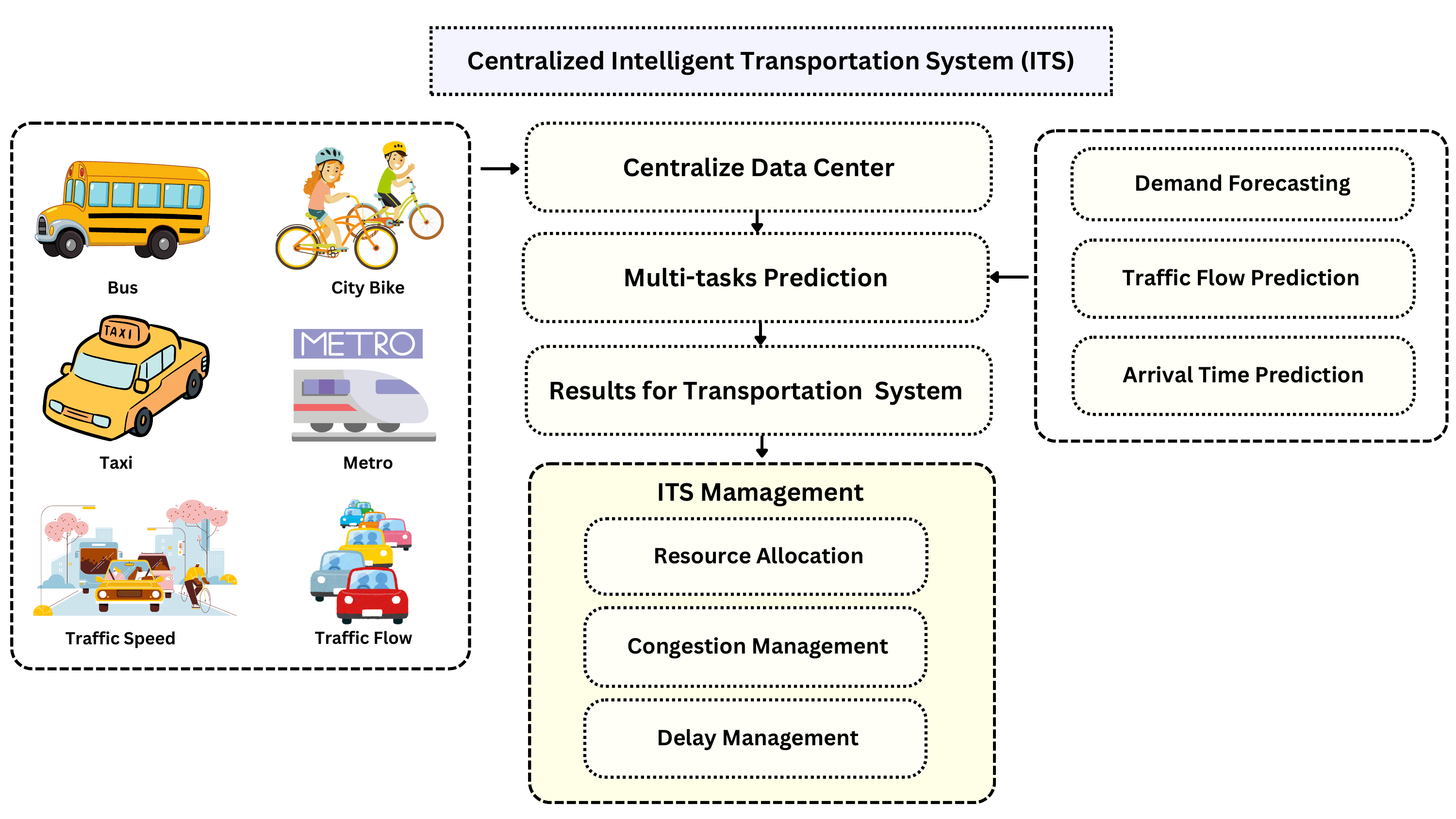}
    \caption{The implementation of an Intelligent Transportation System.}
    \label{fig:intro}
\end{figure}

\section{Preliminary}
\label{Preliminary}
In this section, we briefly introduce the essential notations in Section \ref{Notations}, the fundamental of Denoising Diffusion Probabilistic Models (DDPM) in Section~\ref{ddpm} and the Large Language Model (Llama 2) in Section~\ref{LLM}.

\subsection{Notations and Definition of Problems}
\label{Notations}

\paragraph{Traffic Data} For traffic system data, at any timestamp $t$, it is represented by $\mathbf{X}_t \in \mathbb{R}^{N \times d}$, capturing flow features across $N$ nodes with $d$ denoting the feature dimensions. Over a duration $T>0$, traffic data consolidates into a rank-3 tensor $\mathbf{X} \in \mathbb{R}^{T \times N \times d}$, structuring the temporal, spatial, and feature-specific dimensions of flow. 

\paragraph{Problem Statement}
The goal of traffic flow forecasting is to accurately predict future traffic conditions based on historical data. Consider the traffic flow tensor $\mathbf{X}$ observed in a road network $\mathcal{G}$. We seek to design a function $f$ that utilizes the traffic flow information from the past $m$ timestamps to predict the traffic flow for the next $n$ timestamps. This can be formally expressed as:
\begin{equation*}
f\left( [ \mathbf{X}_{t-m+1}, \ldots,  \mathbf{X}_{t}]  \right) \mapsto [ \mathbf{X}_{t+1}, \ldots,  \mathbf{X}_{t+z}]
\end{equation*}
for any chosen starting point $t$, where $\mathbf{X}_{t}$ represents the traffic flow tensor at time $t$.

\subsection{Denoising Diffusion Probabilistic Models (DDPM)}
\label{ddpm}

Denoising Diffusion Probabilistic Models (DDPMs) represent a novel class of generative models that have shown remarkable proficiency in synthesizing high-fidelity data. These models operationalize a two-phase process: a forward diffusion process that incrementally corrupts the data, followed by a reverse process that aims to reconstruct the original data from the noise.

\paragraph{Forward Diffusion Process}
The forward diffusion process is conceptualized as a Markov chain, where noise is progressively added to the data over a series of discrete time steps. Formally, this process is defined as:
\begin{equation}
\mathbf{X}_{t} = \sqrt{\alpha_t} \mathbf{X}_{t-1} + \sqrt{1 - \alpha_t} \epsilon_t, \quad \epsilon_t \sim \mathcal{N}(0, I),\label{eq:Forward_Diffusion}
\end{equation}
where $ \mathbf{X}_ {t-1}$ denotes the original data, $ \alpha_t $ are pre-defined variance schedules, and $ \epsilon_t $ represents isotropic Gaussian noise. The diffusion process effectively transitions the data distribution towards a Gaussian distribution.

\paragraph{Reverse Diffusion Process}
In contrast, the reverse diffusion process is tasked with learning to invert the forward diffusion. It incrementally denoises the data, guided by a parameterized model $ \epsilon_\theta $. This reverse process is mathematically represented as:
\begin{equation}
\mathbf{X}_{t-1} = \frac{1}{\sqrt{\alpha_t}} \left( \mathbf{X}_{t} - \frac{1 - \alpha_t}{\sqrt{1 - \alpha_t^2}} \epsilon_\theta(\mathbf{X}_{t}, t) \right) + \delta z,\label{eq:Reverse_Diffusion}
\end{equation}
where the function $ \epsilon_\theta(\mathbf{X}_{t}, t) $ approximates the noise added at each diffusion step.

\paragraph{Objective Function}
The primary objective in training DDPMs is to optimize the parameter $ \theta $ such that the model accurately predicts the noise added at each step. This is typically achieved via a mean squared error loss function, defined as:
\begin{equation}
\mathcal{L}(\theta) = \mathbb{E}_{t, \mathbf{X}_0, \epsilon_t} \left[ \| \epsilon_t - \epsilon_\theta(\mathbf{X}_{t}, t) \|^2 \right].
\end{equation}

The optimization of this objective function enables the model to generate new samples from the learned data distribution, effectively reversing the diffusion process.

\subsection{Large Language Models}
\label{LLM}
\paragraph{Information Filtering Hypothesis}
The pre-trained LLM, specifically a transformer-based architecture, operates as an effective ``filter'' in identifying and emphasizing informative tokens. It enhances the significance of these tokens through increased magnitudes or frequencies within the feature activation landscape. This selective amplification facilitates more accurate predictions by foregrounding salient information and effectively streamlining the input data. In this paper, we use Llama 2 \citep{touvron2023llama}, which is built upon the robust Transformer architecture, renowned for its efficacy in handling sequential data through self-attention mechanisms. The core components of Llama 2 include multi-head attention, position-wise feed-forward networks, layer normalization, and residual connections.

\paragraph{Multi-Head Attention Mechanism}

Llama 2 employs multi-head attention to capture different aspects of the input sequence. This mechanism allows the model to focus on various parts of the text simultaneously by using multiple sets of queries, keys, and values. The attention function is defined as:
\begin{equation}
\text{Attention}(Q, K, V) = \text{softmax}\left(\frac{QK^T}{\sqrt{d_k}}\right)V, \label{Eq.4}
\end{equation}
where $ Q $ (queries), $ K $ (keys), and $ V $ (values) are matrices derived from the input data, and $ d_k $ is the dimension of the keys. The multi-head attention mechanism is formulated as:
\begin{align}
  \text{MultiHead}(Q, K, V) &= \text{Concat}(\text{head}_1, \ldots, \text{head}_h)W^O  \\
  \text{head}_i &= \text{Attention}(QW_i^Q, KW_i^K, VW_i^V),\label{Eq.5}
\end{align}
with $ W_i^Q, W_i^K, W_i^V, $ and $ W^O $ being learned projection matrices.

\paragraph{Position-Wise Feed-Forward Networks} Each layer in Llama 2 includes position-wise feed-forward networks that apply two linear transformations with a ReLU activation in between. This is represented as:
\begin{equation}
\text{FFN}(x) = \text{max}(0, xW_1 + b_1)W_2 + b_2,\label{Eq.6}
\end{equation}
where $ W_1 $ and $ W_2 $ are weight matrices, and $ b_1 $ and $ b_2 $ are bias vectors.

\paragraph{Layer Normalization and Residual Connections} Llama 2 incorporates layer normalization and residual connections to stabilize training and improve convergence speed. Layer normalization is defined as:
\begin{equation}
\text{LayerNorm}(x) = \frac{x - \text{E}[x]}{\sqrt{\text{Var}[x] + \epsilon}} \cdot \gamma + \beta,\label{Eq.7}
\end{equation}
where $ \gamma $ and $ \beta $ are learnable parameters. Residual connections help in gradient flow during backpropagation:
\begin{equation}
\text{Output} = \text{LayerNorm}(x + \text{Sublayer}(x))
\end{equation}

\section{Method}
\label{Method}
\begin{figure*}[htbp!]
    \centering
    \includegraphics[scale = 0.3]{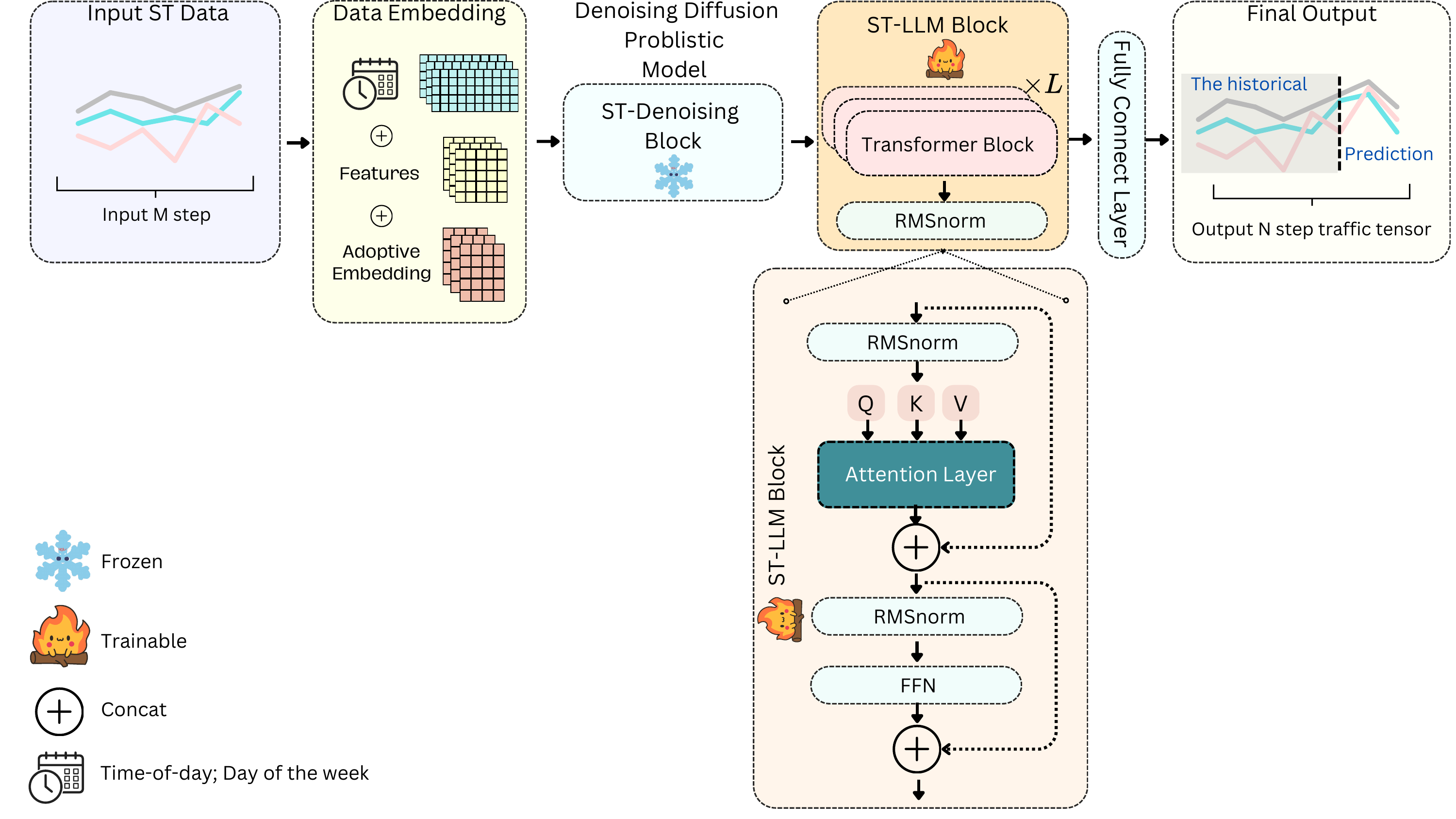}
    \caption{The proposed model framework. }
    \label{fig:framework}
\end{figure*}

In this section, we are going to introduce the proposed model in Section \ref{data_embedding}, \ref{diffusion_frozen}, and \ref{ST-LLM}. The proposed model as in Figure~\ref{fig:framework} integrates various components for effective traffic system forecasting. At its core, the model employs an embedding layer to generate a hidden representation enriched by feature embedding, periodicity embedding, and spatial temporal adaptive embedding. This is followed by the frozen Denoising Diffusion Probalistic Model (DDPM) for initial data recovery, and a trainable layer extracted from Large Language Models (LLMs) serves as an information filter. Then, passing to a fully connected layer tasked with the final traffic prediction.


\subsection{Data Pre-processing}

The data resource for New York City is https://opendata.cityofnewyork.us/. To feed the data into our model, our data processing pipeline was designed to prepare and structure the NYC-Bike, NYC-Bus, NYC-Metro, and NYC-Taxi datasets for time series forecasting. The raw data, originally stored in a pickle file, was first loaded and extracted to obtain the TrafficNode information. We then expanded the dimensionality of the data to accommodate the model's input requirements. Using a custom DataProcessor class, we implemented a sliding window approach to generate input-output pairs, with both input and output windows set to 12 timesteps. This allowed the model to use the past 12 timesteps to predict the next 12 steps. The dataset was then split into training, validation, and test sets with ratios of 70\%, 10\%, and 20\% respectively. To ensure reproducibility and efficient data loading in subsequent runs, we implemented a caching mechanism that saves the processed datasets as compressed NumPy arrays. This preprocessing approach not only structured the data for our forecasting model but also optimized it for efficient training and evaluation processes.

\subsection{Data Embedding Layer}
\label{data_embedding}
We utilized the adoptive data embedding layer as in \cite{staformer}. To preserve the original data's intrinsic information, we apply a fully connected layer to derive the feature embedding $\mathbf{X}_t^f \in \mathbb{R}^{T \times N \times d_f}$:
\begin{equation}
\mathbf{X}_t^f = FC(\mathbf{X}_{t-m+1:t})
\end{equation}
where $d_f$ represents the dimensionality of the feature embedding, and $FC(\cdot)$ denotes a fully connected layer. We introduce the learnable day-of-week embedding dictionary as $\mathbf{T}_w \in \mathbb{R}^{N_w \times d_f}$, and the time-of-day embedding dictionary as $\mathbf{T}_d \in \mathbb{R}^{N_d \times d_f}$, where $N_w = 7$ indicates the week's seven days, and $N_d = 288$ signifies the day's 288 timestamps. Letting $\mathbf{W}_t \in \mathbb{R}^{T}$ and $\mathbf{D}_t \in \mathbb{R}^{T}$ represent the day-of-week and time-of-day data for the traffic series from $t-m+1$ to $t$, we use these as indices to retrieve the respective day-of-week embedding $\mathbf{X}_{t_w} \in \mathbb{R}^{T \times d_f}$ and time-of-day embedding $\mathbf{X}_{t_d} \in \mathbb{R}^{T \times d_f}$. By concatenating and broadcasting these embeddings, we obtain the periodicity embedding $\mathbf{X}_t^p \in \mathbb{R}^{T \times N \times 2d_f}$ for the traffic series.

It's understood that temporal relationships in traffic data are influenced not only by periodicity but also by chronological sequence, where a timeframe should be more similar to its adjacent frames. Moreover, traffic series from different sensors exhibit unique temporal patterns. Thus, instead of using a predefined or dynamic adjacency matrix for spatial relation modeling, we propose a spatio-temporal adaptive embedding $\mathbf{X}_t^a \in \mathbb{R}^{T \times N \times d_a}$ to uniformly capture complex spatio-temporal relationships, shared across different traffic series.

Combining the above embeddings yields the hidden spatio-temporal representation $\mathbf{Z} \in \mathbb{R}^{T \times N \times d_h}$ as:
\begin{equation}
\mathbf{X}_t = \mathbf{X}_t^f \, || \, \mathbf{X}_t^p \, || \, \mathbf{X}_t^a
\end{equation}
where the hidden dimension $d_h$ equals $3d_f + d_a$.

\subsection{Frozen Denoising Block}\label{diffusion_frozen}
Denoising diffusion probabilistic models (DDPM) are commonly used as generative models. In our research, we leverage the data generative capabilities of DDPM to recover the missing traffic data. We have developed a data recovery block using DDPM, explained in Section ~\ref{ddpm}, to effectively remove noise. Compared to traditional denoising/data recovery techniques, the advantage of the diffusion model lies in its strong data recovery capabilities, especially when dealing with data that has complex dependencies. By incorporating time-step embedding, the model can capture the dynamic changes in traffic data over time while using convolutional networks to effectively capture the spatial features of the data. This denoising method, which combines spatio-temporal characteristics, provides a new theoretical perspective and practical approach for processing traffic data. This research extends the theoretical foundation of the diffusion model to the field of traffic data recovery. It not only enriches the theoretical methods for processing traffic data but also provides new insights and frameworks for the application of diffusion models in handling data with complex spatio-temporal dependencies. The Frozen Denoising Block comprises three critical components: time step embedding, a feature extraction network, and a data recovery network. These elements are elaborated in the subsequent paragraphs:

\paragraph{Time Step Embedding}
The timestep embedding function is defined to map a 1-D tensor of time indices $ t $ into a $ N \times \text{dim} $ tensor of positional embeddings, where each embedding captures periodic patterns across different frequencies. This is formulated as:

\begin{equation}
\textbf{emb}(t, d) = 
\begin{cases} 
\cos\left( \frac{t \times \exp\left(-\frac{k \log(\text{max\_period})}{\text{dim}/2}\right)}{\text{max\_period}} \right) & \text{if } 0 \leq k < \text{dim}/2 ,\\
\sin\left( \frac{t \times \exp\left(-\frac{(k - \text{dim}/2) \log(\text{max\_period})}{\text{dim}/2}\right)}{\text{max\_period}} \right) & \text{if } \text{dim}/2 \leq k < \text{dim}.
\end{cases}
\end{equation}

Where $ t $ is the time index, which can be fractional. $ \text{dim} $ is the dimensionality of the output embedding. $ \text{max\_period} $ controls the maximum period of the sinusoidal functions, influencing the minimum frequency. $ k $ represents the dimension index within the embedding vector.

The embeddings for each dimension $ k $ are derived from the exponential spacing of frequencies, ensuring coverage across a range of scales from the most fine-grained changes at the highest frequency to the most coarse-grained changes at the lowest frequency. The function ensures that each time step $ t $ is mapped distinctly, capturing periodic recurrences in data effectively.

\paragraph{Feature Extraction Network}
For the feature extraction, we simply Leverage the power of convolutional networks to extract crucial spatial features. By integrating time step embedding, our network adeptly captures the dynamic variations in traffic data over time.
\begin{align}
     \mathbf{X}_t &=\textbf{Conv} \Big( \textbf{Conv}(\mathbf{X}_t) + \textbf{emb}(t, d)  \Big)
\end{align}
\paragraph{Data Recovery Network}
Based on the results of time step embedding and feature extraction, the data recovery network learns to predict noise and gradually reconstructs clean data through the reverse diffusion process.

The forward step of DDPM for adding noise, described in Eq.~\ref{eq:Forward_Diffusion}, is formulated as:
\begin{equation}
    \mathbf{X}_{t+1} = \sqrt{\alpha_t} \mathbf{X}_{t} + \sqrt{1 - \alpha_t} \epsilon_t, \quad \epsilon_t \sim \mathcal{N}(0, I),
\end{equation}
here, the $\mathbf{X}_{t}$ is the input traffic data at timestep $t$, $\mathbf{X}_{t+1}$ is the generated traffic data $\epsilon_t$ is Gaussian noise with standard deviation $I$, $\alpha_t$ is a coefficient that gradually decreases over time.
The translation for the text is:

Since the forward process is known, we can train a network to predict the noise given $ x_t $, thereby gradually reconstructing $ x_0 $. The reverse diffusion process, which aids in handling missing data, is formulated as in Eq.~\ref{eq:Reverse_Diffusion} with the following:
\begin{equation}
\mathbf{X}_{t} = \frac{1}{\sqrt{\alpha_t}} \left( \mathbf{X}_{t+1} - \frac{1 - \alpha_t}{\sqrt{1 - \alpha_t^2}} \epsilon_\theta(\mathbf{X}_{t}, t) \right) + \delta z,
\end{equation}
where $\delta z$ is a residual term added to refine the denoising output.

\subsection{Spatial Temporal Large Language Model Block}\label{ST-LLM}
Current traffic models, especially CNNs and RNNs, face challenges in capturing both spatial and temporal dependencies, while large language models excel at time series analysis but often need to pay more attention to the spatial aspect of traffic prediction tasks. To address these issues, a novel spatial temporal large-scale language model block (ST-LLM) has been proposed, 
the formulation as in Eq.~\ref{Eq.4} - ~\ref{Eq.7}:
\begin{align}
    \mathbf{\Tilde{X}}_{t} &= \text{RMSNorm} (\text{MultiHead}(Q, K, V)) + \mathbf{X}_{t},\\
    \mathbf{\Bar{X}}_{t} & = \text{FFN} \Big( \text{RMSNorm}(\mathbf{\Tilde{X}}_{t}) \Big) +  \mathbf{\Tilde{X}}_{t}.
\end{align}
This block redefines the time step of each location as a marker and incorporates a spatial temporal embedding module to learn the markers' spatial position and global time representation. It aims to effectively capture both spatial and temporal dependencies in traffic data, providing a practical and relevant solution for transportation and urban planning professionals.

The Final output can be obtained by a fully connected layer as:
\begin{equation}
    \mathbf{Y}_t = FC(\mathbf{\Bar{X}}_{t}),
\end{equation}
where $\mathbf{Y}_t \in \mathbb{R}^{z \times N \times d}$.

\begin{algorithm}[H]
\SetAlgoLined
\KwIn{Embedded data $\mathbf{Z}$}
\KwOut{Prediction of $\mathbf{X}_{t+1:t+n}$}

\textbf{Step 1: Embedding Layer}\;
Obtain feature embedding $\mathbf{X}_t^f$ from $\mathbf{X}_{t-m+1:t}$\;
$\mathbf{Z} \leftarrow \mathbf{X}_t^f \, || \, \mathbf{X}_t^p \, || \, \mathbf{X}_t^a$\;

\textbf{Step 2: Denoising with DDPM}\;
\For{$i \leftarrow t-m+1$ \KwTo $t$}{
    Recover $\mathbf{X}_i$ using DDPM to obtain $\mathbf{\bar{X}}_i$\;
}

\textbf{Step 3: Information Filtering with Frozen LLM}\;
Filter $\mathbf{\bar{X}}_{t-m+1:t}$ through frozen LLM to obtain $\mathbf{Y}_{t-m+1:t}$\;

\textbf{Step 4: Temporal and Spatial Transformer Layers}\;
Apply Temporal Transformer on $\mathbf{Y}_{t-m+1:t}$ to capture temporal dynamics\;
Apply Spatial Transformer on output of Temporal Transformer to capture spatial relations\;

\textbf{Step 5: Fully Connected Layer for Prediction}\;
Predict $\mathbf{X}_{t+1:t+n}$ using the regression layer\;

\caption{STLLM-DF: Traffic Flow Prediction Model}
\end{algorithm}

\section{Experiment}
\label{Experiment}
In this section, we provide the dataset description in Section~\ref{Dataset}, followed by model configurations and evaluation metrics in Section~\ref{Model configurations and evaluation metrics}. Baseline models are discussed in Section~\ref{Baseline}, while model performance is evaluated in Section~\ref{Model Performance}. An ablation study is presented in Section~\ref{Ablation Study}, concluding with a real case analysis in Section~\ref{Real Case Analysis}.

\subsection{Dataset Description}
\label{Dataset}
In our study, we evaluate STLLM-DF using six real-world datasets in both graph-based and grid-based data structures. Specifically, the graph-based datasets include \texttt{PeMS04}, \texttt{PeMS07}, and \texttt{PeMS08}, while the grid-based datasets comprise \texttt{CHIBike}, \texttt{TDrive}. To test the handling multi-task ability for STLLM-DF , we further add New York City Transportation Datasets including \texttt{NYC-BIKE}, \texttt{NYC-BUS}, \texttt{NYC-TAXI} and \texttt{NYC-METRO} . These datasets are publicly accessible and can be found in the LibCity repository, as referenced in \cite{wang2023libcity}. Detailed information about each of these datasets is provided in Table \ref{table:data11} and \ref{table:data2}.
\begin{table}[htbp!]
    \centering
    \footnotesize
    \caption{Statistics of traffic flow forecasting datasets.}
    \begin{tabular}{@{}lcccccc@{}}
        \toprule
        \textbf{Dataset} & \textbf{PEMS03} & \textbf{PEMS04} & \textbf{PEMS07} & \textbf{PEMS08} & \textbf{METR-LA} & \textbf{PeMS-BAY} \\
        \midrule
        \# of nodes       & 358      & 307      & 883      & 170      & 207      & 325      \\
        \# of timesteps   & 26,208   & 16,992   & 28,224   & 17,856   & 34,272   & 52,116   \\
        Granularity       & 5min     & 5min     & 5min     & 5min     & 5min     & 5min     \\
        Start time        & 9/1/2018 & 1/1/2018 & 5/1/2017 & 7/1/2016 & 3/1/2012 & 1/1/2017 \\
        End time          & 11/30/2018 & 2/28/2018 & 8/31/2017 & 8/31/2016 & 6/30/2012 & 5/31/2017 \\
        Missing ratio     & 0.672\%  & 3.182\%  & 0.452\%  & 0.696\%  & 8.11\%   & 0.003\%  \\
        \# Signals        & F        & F, S, O  & F        & F, S, O  & S        & S        \\
        \bottomrule
    \end{tabular}
    \label{table:data11}
\end{table}

\begin{table}[htbp!]
    \centering
    \footnotesize
    \caption{Statistics of New York City transportation datasets.}
    \begin{tabular}{@{}lccccc@{}}
        \toprule
        \textbf{Dataset} & \textbf{NYC-BIKE} & \textbf{NYC-BUS} & \textbf{NYC-TAXI} & \textbf{NYC-METRO} \\
        \midrule
        \# of nodes       & 820      & 226      & 263      & 426      \\
        \# of timesteps   & 446976 (5 mins)  & 17,064 & 1,516,032 & 16,512 \\
        Granularity       & 5  & 60 mins & 5 mins & 60 mins \\
        Start time        & 7/1/2013    & 2/1/2022 & 1/1/2009 & 2/1/2022 \\
        End time          & 9/30/2017 & 1/13/2024 & 6/1/2023 & 12/21/2023 \\
         Missing ratio     & 1.232\% & \\
        \# Signals        & D        & F  & D       & D        \\
        \bottomrule
    \end{tabular}
    \label{table:data2}
\end{table}

\textbf{\texttt{PeMS04} \cite{song2020spatial}:} Representing traffic data from the San Francisco Bay Area, this dataset was accumulated by the Caltrans Performance Measurement Systems (PeMS). Data from one sensor is condensed into 5-minute intervals, incorporating traffic flow, average speed, and average occupancy. It encompasses records from 307 sensors, spanning from Jan 1, 2018, to Feb 28, 2018.

\textbf{\texttt{PeMS07} \cite{song2020spatial}:} Representing traffic data from the San Francisco Bay Area, this dataset was accumulated by the Caltrans Performance Measurement Systems (PeMS). Data from one sensor is condensed into 5-minute intervals, incorporating traffic flow, average speed, and average occupancy. It encompasses records from 883 sensors, spanning from May 1, 2017, to Aug 31, 2017.

\textbf{\texttt{PeMS08} \cite{song2020spatial}:} This is the highway traffic flow data collected by the California Department of Transportation (Caltrans). The
data range is from Jul 1, 2016 to Aug 31, 2016. The flow data is sampled every 5 minutes. The number of sensors is 170. 

\begin{figure} [htbp!]
    \centering
    \includegraphics[scale = 0.3]{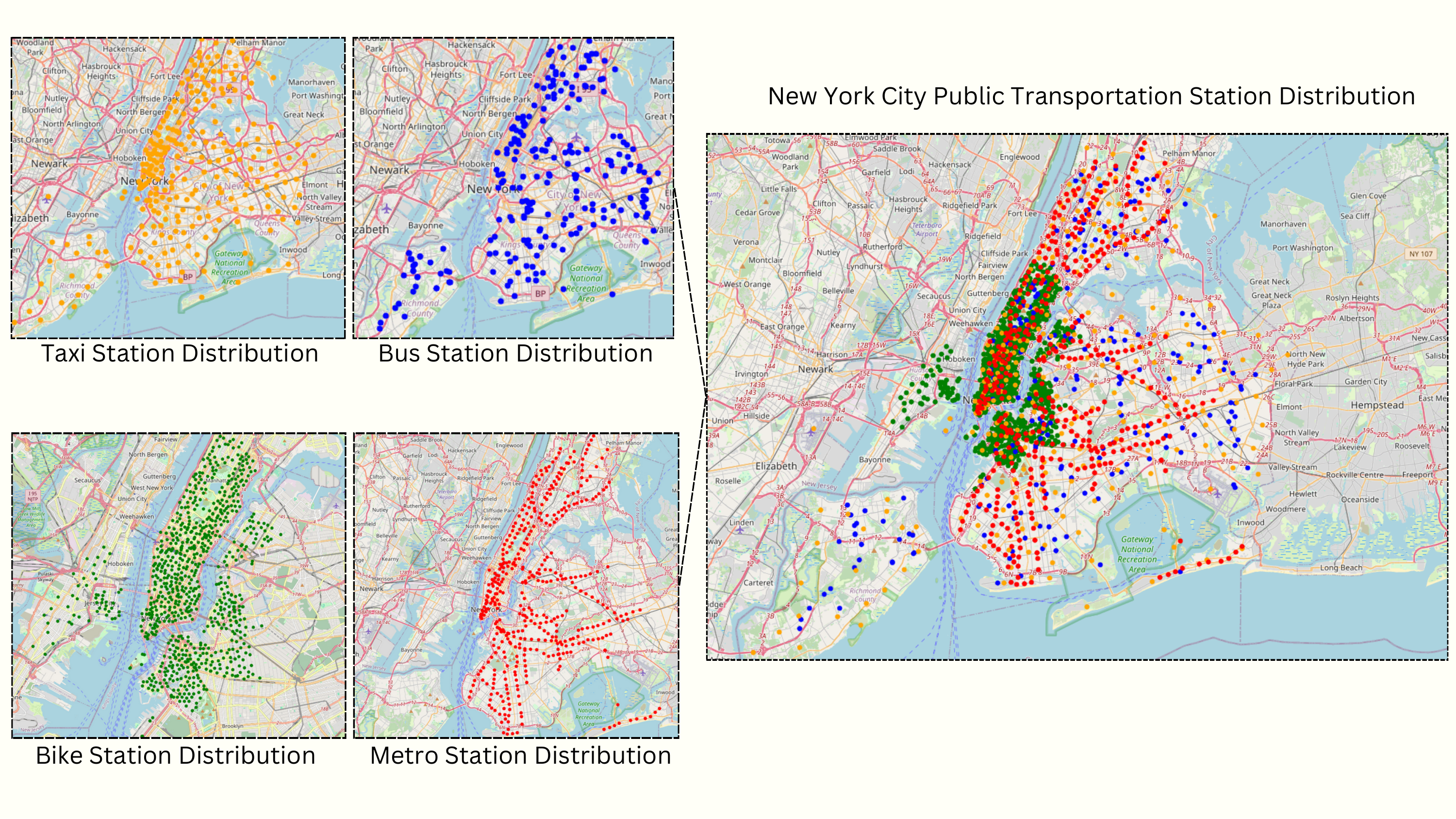}
    \caption{Public transportation station distribution in New York City.}
    \label{fig:distribution}
\end{figure}

\textbf{\texttt{NYC-BIKE}:} The NYC Bike Sharing Dataset provides a comprehensive view of New York City's bike-sharing system with hourly granularity from July 1, 2013 to September 30, 2017. It encompasses multiple interconnected components: TrafficNode (446976x820x820) capturing hourly trips between 820 stations, TrafficMonthlyInteraction (51x820x820) for monthly aggregated trips, StationInfo (820x5) detailing station locations, TrafficGrid (446976x20x20) representing city-wide usage patterns, weather data (446976x24), and social media check-in features. This rich dataset enables diverse analyses, including predictive modeling of bike demand, route popularity studies, weather impact assessments, and urban mobility pattern examinations, the station distribution as showing in Figure~\ref{fig:distribution}.

\textbf{\texttt{NYC-BUS}:} The NYC Bus Dataset provides a comprehensive view of New York City's bus system from February 1, 2022 to January 13, 2024, offering hourly granularity data for 226 bus nodes or stations. The core of the dataset is a TrafficNode array (17064x226x226) capturing hourly passenger flow between all pairs of bus stops. It includes detailed StationInfo for each stop, comprising ID, geographical coordinates, and stop names. The dataset features 17,064 hourly time slots over nearly two years, allowing for in-depth analysis of daily, weekly, and seasonal trends in bus ridership. Additional components include a TrafficGrid representing bus traffic in a 20x20 city grid, weather data with 24 state dimensions, and potential social media check-in features, the station distribution as showing in Figure~\ref{fig:distribution}.

\textbf{\texttt{NYC-taxi} \cite{liu2020dynamic}:} The data was made available by the \texttt{NYCtaxi} \& Limousine Commission (TLC) and built on data from ride-hailing companies. The records cover New York from Jan 1, 2014 to Dec 31, 2014. For each demand record, the data provides information such as the pick-up time, drop-off time, pick-up zone, drop-off zone, etc. The traffic zones are predetermined by
the TLC. In our dataset, we let each NYC taxi zones as a node, each node represents a specific area in NYC where taxis pick up or drop off passengers, the station distribution as showing in Figure~\ref{fig:distribution}.

\textbf{\texttt{NYC-METRO}:} The NYC Metro Dataset offers a comprehensive view of New York City's subway system from February 1, 2022 to December 21, 2023, with hourly granularity. It features a multi-dimensional TrafficNode array (16512x472x472) capturing passenger flow between 472 subway stations and detailed StationInfo for each station, including geographical coordinates and names. The dataset's core is a 3D matrix where each slice represents hourly passenger movements between all station pairs, allowing for in-depth analysis of travel patterns and the station distribution as shown in Figure~\ref{fig:distribution}.

\subsection{Model Configurations and Evaluation Metrics}
\label{Model configurations and evaluation metrics}
\paragraph{Model configurations} All experiments were executed on a machine featuring an A100-SXM4-80GB. The datasets PEMS-BAY, PEMS03, PEMS04, PEMS07, PEMS08 and NYC datasets were split into training, validation, and test sets. PEMS-BAY and NYC datasets were divided in a 7:1:2 ratio, while PEMS03, PEMS04, PEMS07, and PEMS08 were partitioned using a 6:2:2 ratio. ST-LLM (Llamma) block-based configuration was also implemented with 32 heads, 17 layers, and a normalization epsilon of 1.0e-6. Notably, the model incorporated an ST-Denoising block (DDIMSampler) for the data recovery process, configured with a beta range of [0.0001, 0.02] and 1000 sampling steps, enhancing the model's capability to generate refined predictions through iterative data recovery.

\paragraph{Evaluation metrics}
To evaluate the effectiveness of traffic forecasting methods, three commonly used metrics are applied: \textit{Mean Absolute Error} (MAE), \textit{Mean Absolute Percentage Error} (MAPE), and \textit{Root Mean Square Error} (RMSE). These metrics provide a comprehensive assessment of model accuracy and the magnitude of errors. They are defined as follows:

\textbf{Mean Absolute Error (MAE)} measures the average magnitude of prediction errors without considering their direction. It is calculated using the formula:
\[
\text{MAE} = \frac{1}{n} \sum_{i=1}^{n} |\hat{y}_i - y_i|,
\]
where \( n \) represents the number of observations, \( y_i \) denotes the actual values, and \( \hat{y}_i \) represents the predicted values.

\textbf{Mean Absolute Percentage Error (MAPE)} expresses the error as a percentage of the actual values, allowing for easy comparison across datasets with different scales. It is computed as:
\[
\text{MAPE} = \frac{1}{n} \sum_{i=1}^{n} \left| \frac{\hat{y}_i - y_i}{y_i} \right| \times 100,
\]
where \( y_i \) and \( \hat{y}_i \) are the actual and predicted values, respectively.

\textbf{Root Mean Square Error (RMSE)} calculates the square root of the average squared differences between predicted and actual values, placing greater emphasis on larger errors. The formula is:
\[
\text{RMSE} = \sqrt{\frac{1}{n} \sum_{i=1}^{n} (\hat{y}_i - y_i)^2}.
\]

Employing MAE, MAPE, and RMSE provides a thorough analysis of model performance in traffic forecasting, capturing not only the average errors but also the distribution and proportionality of these errors relative to the actual values.

\subsection{Baseline Models}
\label{Baseline}
In our comparative analysis, we evaluate our proposed approach against a comprehensive set of baselines in the traffic forecasting domain:

\begin{itemize}
    \item \textbf{Historical Index (HI)} \cite{cui2021historical}: Serves as the conventional benchmark, reflecting standard industry practices.
    
    \item \textbf{Spatial-Temporal Graph Neural Networks (STGNNs)}:
    \begin{itemize}
        \item \textbf{GWNet} \cite{wu2020connecting}: Automatically extracts uni-directed relations among variables, addressing limitations in exploiting latent spatial dependencies.
        
        \item \textbf{DCRNN} \cite{li2017diffusion}: Introduces the Diffusion Convolutional Recurrent Neural Network, capturing both spatial and temporal dependencies.
        
        \item \textbf{AGCRN} \cite{bai2020adaptive}: Incorporates adaptive modules to capture node-specific patterns and infer inter-dependencies among traffic series.
        
        \item \textbf{STGCN} \cite{ijcai2018p505}: Integrates graph convolutions for spatial feature extraction and gated temporal convolutions for temporal feature extraction.
        
        \item \textbf{GTS} \cite{shang2021discrete}: Forecasts multiple interrelated time series by learning a graph structure simultaneously with a Graph Neural Network.
        
        \item \textbf{MTGNN} \cite{Wu2020ConnectingTD}: Automatically extracts uni-directed relations among variables, capturing both spatial and temporal dependencies.
        
        \item \textbf{GMAN} \cite{Zheng_Fan_Wang_Qi_2020}: Incorporates spatial and temporal attention mechanisms to capture dynamic correlations among traffic sensors.
    \end{itemize}
    
    \item \textbf{Transformer-based models}:
    \begin{itemize}
        \item \textbf{PDFormer} \cite{pdformer}: Captures dynamic spatial dependencies, long-range spatial dependencies, and time delay in traffic condition propagation.
        
        \item \textbf{STAEformer} \cite{staeformer}: Proposes a spatio-temporal adaptive embedding that enhances the performance of vanilla transformers.
        
        \item \textbf{STNorm} \cite{dengstnorm}: Leverages spatial and temporal normalization modules to refine high-frequency and local components.
        
        \item \textbf{STID} \cite{shaozhao2022}: Addresses indistinguishability of samples by attaching spatial and temporal identity information to input data.
    \end{itemize}
\end{itemize}

This diverse range of models allows for a robust validation of our proposed method's capabilities in traffic forecasting.

\subsection{Model Performance}
\label{Model Performance}

\begin{table}[htbp!]
\centering
\caption{Performance comparison of models on PEMS datasets.}
\label{tab:performance_comparison_whole}
\setlength{\tabcolsep}{3pt}
\resizebox{\textwidth}{!}{%
\begin{tabular}{l S[table-format=2.2] S[table-format=2.2] c S[table-format=2.2] S[table-format=2.2] c S[table-format=2.2] S[table-format=2.2] c S[table-format=2.2] S[table-format=2.2] c}
\toprule
{Model} & \multicolumn{3}{c}{PEMS03} & \multicolumn{3}{c}{PEMS04} & \multicolumn{3}{c}{PEMS07} & \multicolumn{3}{c}{PEMS08} \\
\cmidrule(lr){2-4} \cmidrule(lr){5-7} \cmidrule(lr){8-10} \cmidrule(lr){11-13}
& {MAE} & {RMSE} & {MAPE} & {MAE} & {RMSE} & {MAPE} & {MAE} & {RMSE} & {MAPE} & {MAE} & {RMSE} & {MAPE} \\
\midrule
HI & 32.62 & 49.89 & 30.60\% & 42.35 & 61.66 & 29.92\% & 49.03 & 71.18 & 22.75\% & 36.66 & 50.45 & 21.63\% \\
GWNet & \textcolor{red}{\textbf{14.59 }}& \textcolor{red}{\textbf{25.24}} & 15.52\% & 18.53 & 29.92 & 12.89\% & 20.47 & 33.47 & 8.61\% & 14.40 & 23.39 & 9.21\% \\
DCRNN & 15.54 & 27.18 & 15.62\% & 19.63 & 31.26 & 13.59\% & 21.16 & 34.14 & 9.02\% & 15.22 & 24.17 & 10.21\% \\
AGCRN & 15.24 & 26.65 & 15.89\% & 19.38 & 31.25 & 13.40\% & 20.57 & 34.40 & 8.74\% & 15.32 & 24.41 & 10.03\% \\
STGCN & 15.83 & 27.51 & 16.13\% & 19.57 & 31.38 & 13.44\% & 21.74 & 35.27 & 9.24\% & 16.08 & 25.39 & 10.60\% \\
GTS & 15.41 & 26.15 & 15.39\% & 20.96 & 32.95 & 14.66\% & 22.15 & 35.10 & 9.38\% & 16.49 & 26.08 & 10.54\% \\
MTGNN & 14.85 & 25.23 & 14.55\% & 19.17 & 31.70 & 13.37\% & 20.89 & 34.06 & 9.00\% & 15.18 & 24.24 & 10.20\% \\
STNorm & 15.32 & 25.93 & 14.37\% & 18.96 & 30.98 & 12.69\% & 20.50 & 34.66 & 8.75\% & 15.41 & 24.77 & 9.76\% \\
GMAN & 16.87 & 27.92 & 18.23\% & 19.14 & 31.60 & 13.19\% & 20.97 & 34.10 & 9.05\% & 15.31 & 24.92 & 10.13\% \\
PDFormer & 14.94 & 25.39 & 15.82\% & 18.36 & 30.03 & 12.00\% & 19.97 & 32.95 & 8.55\% & 13.58 & 23.41 & 9.05\% \\
STID & 15.33 & 27.40 & 16.40\% & 18.38 & 29.95 & 12.04\% & 19.61 & 32.79 & 8.30\% & 14.21 & 23.28 & 9.27\% \\
STAEformer& 15.35 & 27.55 & 15.18\% & 18.22 & 30.18 & 11.98\% & 19.14 & 32.60 & 8.02\% & 13.46 & 23.25 & 8.88\% \\
\hline
STLLM-DF  & 15.28 & 27.28& \textcolor{red}{\textbf{15.02\%}} & \textcolor{red}{\textbf{18.15}} &\textcolor{red}{\textbf{29.95}} & \textcolor{red}{\textbf{11.88\%}} &\textcolor{red}{\textbf{19.10}}& \textcolor{red}{\textbf{32.43}}& \textcolor{red}{\textbf{8.01\%}} & \textcolor{red}{\textbf{13.38}} & \textcolor{red}{\textbf{23.08}} & \textcolor{red}{\textbf{8.79\%}}  \\
\bottomrule
\end{tabular}
} 
\end{table}

\paragraph{Analysis of Multi-Task Prediction Results}
\begin{table}[htbp!]
\centering
\caption{Performance comparison of models on NYC transportation datasets.}
\label{tab:nyc_performance_comparison}
\resizebox{\textwidth}{!}{%
\begin{tabular}{lcccccccccccc}
\toprule
\textbf{Model} & \multicolumn{3}{c}{NYC-BIKE} & \multicolumn{3}{c}{NYC-BUS} & \multicolumn{3}{c}{NYC-TAXI} & \multicolumn{3}{c}{NYC-METRO} \\
\cmidrule(lr){2-4} \cmidrule(lr){5-7} \cmidrule(lr){8-10} \cmidrule(lr){11-13}
& {MAE} & {RMSE} & {MAPE} & {MAE} & {RMSE} & {MAPE} & {MAE} & {RMSE} & {MAPE} & {MAE} & {RMSE} & {MAPE} \\
\midrule
STID & 5.01 & 2.70 & 63.54\% & 50.03 & 27.21 & 67.32\% & 3.57 & 2.31 & 49.52\% & 39.04 & 113.28& 60.12\%\\
STAFormer & 5.02 & 2.73 & \textcolor{red}{\textbf{63.43\%}} & 49.01 & 27.64 & 65.2\% & 3.53 & 2.42 & 49.25\%  & 39.11 & 114.12& 60.31\% \\
\hline
STLLM-DF & \textcolor{red}{\textbf{4.97}} & \textcolor{red}{\textbf{2.64}} & 63.44\% & \textcolor{red}{\textbf{47.92}} & \textcolor{red}{\textbf{26.78}} & \textcolor{red}{\textbf{64.43\%}} & \textcolor{red}{\textbf{3.35}} & \textcolor{red}{\textbf{2.19}} & \textcolor{red}{\textbf{47.38\%}} & \textcolor{red}{\textbf{38.30}} & \textcolor{red}{\textbf{112.68}} & \textcolor{red}{\textbf{59.67\%}} \\

\bottomrule
\end{tabular}
}
\end{table}

Table~\ref{tab:nyc_performance_comparison} presents a comparative analysis of two models, STLLM-DF and STAFormer, across four distinct NYC transportation datasets: BIKE, BUS, TAXI, and METRO. The performance metrics utilized for this comparison are Mean Absolute Error (MAE), Root Mean Square Error (RMSE), and Mean Absolute Percentage Error (MAPE).  On average, STLLM-DF achieves improvements of 2.40\% in MAE, 4.50\% in RMSE, and 1.51\% in MAPE. The NYC-TAXI dataset demonstrates the most substantial enhancements, with reductions of 5.10\% in MAE, 9.50\% in RMSE, and 3.80\% in MAPE, indicating STLLM-DF's particular effectiveness in modeling complex transportation systems. 

\begin{figure}[htbp!]
    \centering
    \includegraphics[scale = 0.25]{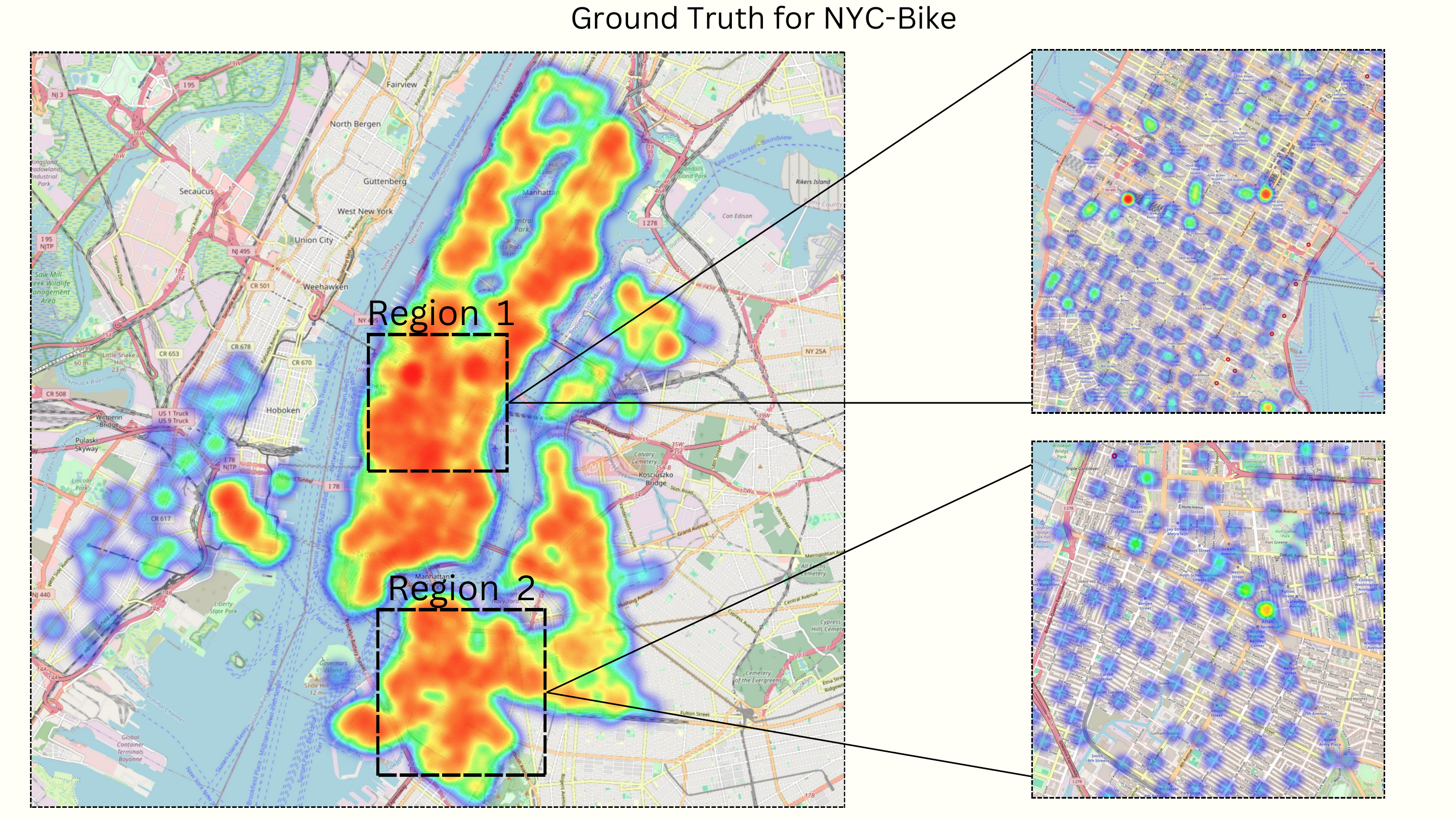}
    \caption{Heatmap visualization of ground truth bike demand data for the NYC-BIKE dataset. }
    \label{fig:heat_nycbike_true}
\end{figure}

\begin{figure}[htbp!]
    \centering
    \includegraphics[scale = 0.25]{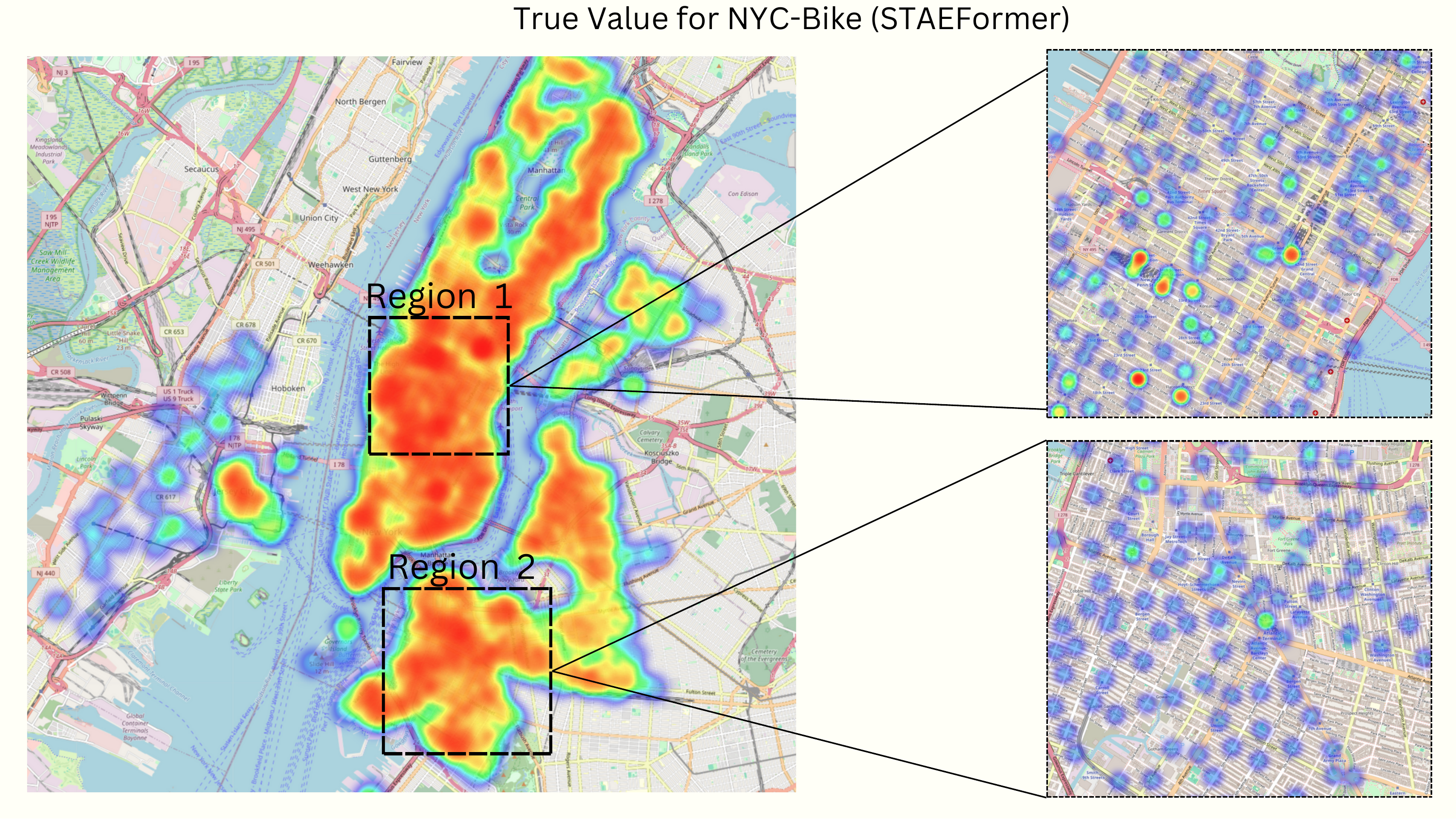}
    \caption{Heatmap visualization of STAEformer bike demand data for the NYC-BIKE dataset. .}
    \label{fig:heat_nycbike_staeformer}
\end{figure}

\begin{figure}[htbp!]
    \centering
    \includegraphics[scale = 0.25]{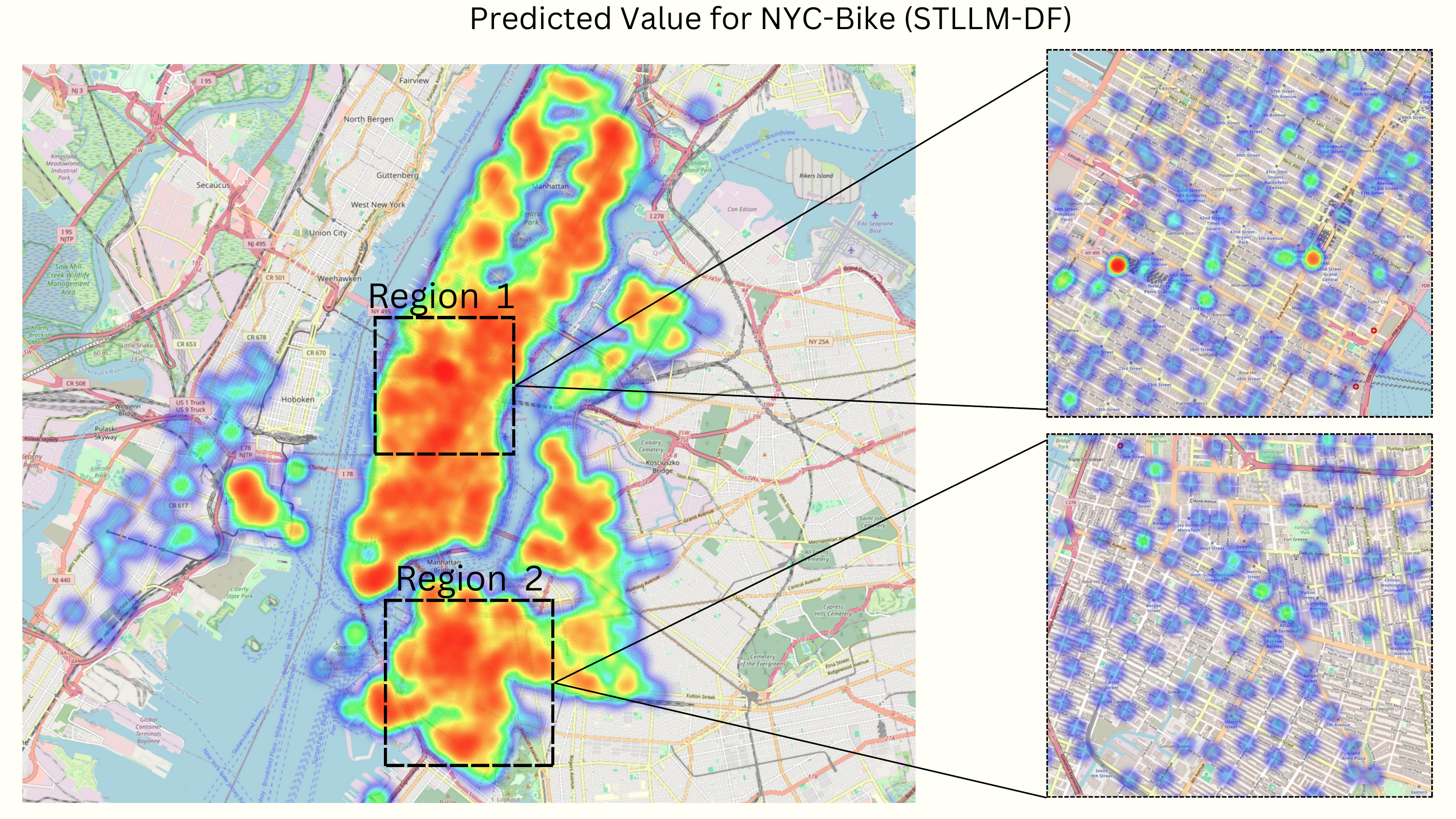}
    \caption{Heatmap visualization of STLLM-DF bike demand data for the NYC-BIKE dataset. }
    \label{fig:heat_nycbike_stllmdf}
\end{figure}

This improvement is further highlighted in the visual comparisons shown in Figures \ref{fig:heat_nycbike_true}, \ref{fig:heat_nycbike_staeformer} and \ref{fig:heat_nycbike_stllmdf}. Figure \ref{fig:heat_nycbike_staeformer} displays the predicted bike demand generated by the STAFormer model, capturing certain regions where the model closely follows the true values. In contrast, Figure \ref{fig:heat_nycbike_stllmdf} demonstrates the predicted bike demand from the STLLM-DF model, which showcases enhanced accuracy, particularly in regions with high demand variability. These visualizations clearly depict the effectiveness of STLLM-DF in capturing the peaks and fluctuations across different areas, notably in ``Region 1'' and ``Region 2''. The detailed insets provide a closer look at localized demand, with STLLM-DF demonstrating a stronger alignment with the ground truth, reinforcing the quantitative performance improvements seen in the table.

\begin{table}[htbp!]
\centering
\caption{Performance on METR-LA and PEMS-BAY.}
\setlength{\tabcolsep}{0.8pt} 
\resizebox{\textwidth}{!}{%
\begin{tabular}{llcccccccccccccc}
\hline
 Horizon & Metric & HI & GWNet & DCRNN & AGCRN & STGCN & GTS & MTGNN & STNorm & GMAN & PDFormer & STID & STAEformer & STLLM-DF\\
\hline
\hline
\multicolumn{14}{c}{METR-LA} \\
\hline
15 min & MAE & 6.80 & 2.69 & 2.67 & 2.85 & 2.75 & 2.75 & 2.69 & 2.81 & 2.80 & 2.83 & 2.82 & 2.65 & \textcolor{red}{\textbf{2.61}}\\
& RMSE & 14.21 & 5.15 & 5.16 & 5.53 & 5.29 & 5.27 & 5.16 & 5.57 & 5.55 & 5.45 & 5.53 & 5.11& \textcolor{red}{\textbf{5.10}} \\
& MAPE & 16.72 & 6.99 & 6.86 & 7.63 & 7.10 & 7.12 & 6.89 & 7.40 & 7.41 & 7.77 & 7.75 & 6.85 & \textcolor{red}{\textbf{6.71}}\\
30 min & MAE & 6.80 & 3.08 & 3.12 & 3.20 & 3.15 & 3.14 & 3.05 & 3.18 & 3.12 & 3.20 & 3.19 & 2.97 & \textcolor{red}{\textbf{2.90}}\\
& RMSE & 14.21 & 6.20 & 6.27 & 6.52 & 6.35 & 6.33 & 6.13 & 6.59 & 6.49 & 6.46 & 6.57 & \textcolor{red}{\textbf{6.00}} & 6.01\\
& MAPE & 16.72 & 8.47 & 8.42 & 9.00 & 8.62 & 8.62 & 8.16 & 8.47 & 8.73 & 9.19 & 9.39 & 8.13 & \textcolor{red}{\textbf{8.06}}\\
60 min & MAE & 6.80 & 3.51 & 3.54 & 3.59 & 3.60 & 3.59 & 3.47 & 3.57 & 3.44 & 3.62 & 3.55 & 3.34 & \textcolor{red}{\textbf{3.27}}\\
& RMSE & 14.20 & 7.28 & 7.47 & 7.45 & 7.43 & 7.44 & 7.21 & 7.51 & 7.35 & 7.47 & 7.55 & \textcolor{red}{\textbf{7.02}} & 7.00\\
& MAPE & 10.15 & 9.96 & 10.32 & 10.47 & 10.35 & 10.25 & 9.70 & 10.24 & 10.07 & 10.91 & 10.95 & 9.70 & \textcolor{red}{\textbf{9.46}}\\
\hline
\multicolumn{14}{c}{PEMS-BAY} \\
\hline
15 min & MAE & 3.06 & 1.30 & 1.31 & 1.35 & 1.36 & 1.37 & 1.33 & 1.33 & 1.35 & 1.32 & 1.31 & 1.31 & \textcolor{red}{\textbf{1.28}}\\
& RMSE & 7.05 & 2.73 & 2.76 & 2.88 & 2.88 & 2.92 & 2.80 & 2.82 & 2.90 & 2.83 & 2.79 & 2.78 & \textcolor{red}{\textbf{2.68}}\\
& MAPE & 6.85 & 2.71 & 2.73 & 2.91 & 2.86 & 2.85 & 2.81 & 2.76 & 2.87 & 2.78 & 2.78 & 2.76 & \textcolor{red}{\textbf{2.70}}\\
30 min & MAE & 3.06 & 1.63 & 1.65 & 1.67 & 1.70 & 1.72 & 1.66 & 1.65 & 1.65 & 1.64 & 1.64 & 1.62 & \textcolor{red}{\textbf{1.60}}\\
& RMSE & 7.04 & 3.73 & 3.75 & 3.82 & 3.84 & 3.86 & 3.77 & 3.77 & 3.82 & 3.79 & 3.73 & \textcolor{red}{\textbf{3.68}} & 3.70\\
& MAPE & 6.84 & 3.73 & 3.71 & 3.81 & 3.79 & 3.88 & 3.75 & 3.66 & 3.74 & 3.71 & 3.73 & 3.62 & \textcolor{red}{\textbf{3.54}}\\
60 min & MAE & 3.05 & 1.99 & 1.97 & 1.94 & 2.02 & 2.06 & 1.95 & 1.92 & 1.91 & 1.91 & 1.91 & 1.88 & \textcolor{red}{\textbf{1.82}}\\
& RMSE & 7.03 & 4.60 & 4.60 & 4.50 & 4.63 & 4.60 & 4.50 & 4.45 & 4.49 & 4.43 & 4.42 & \textcolor{red}{\textbf{4.34}} & 4.37\\
& MAPE & 6.83 & 4.71 & 4.68 & 4.55 & 4.72 & 4.88 & 4.62 & 4.46 & 4.52 & 4.51 & 4.55 & 4.41 & \textcolor{red}{\textbf{4.40}}\\
\hline
\end{tabular}
} 
\label{tab:horizon}
\end{table}

\subsection{Ablation Study and Denoising Analysis}
\label{Ablation Study}
\begin{table}[htbp!]
\centering
\caption{Performance metrics for various models on the PEMS08 dataset.}

\begin{tabular}{@{}lcccccc@{}}
\toprule
Model                    & MAE   & RMSE  & MAPE  \\ \midrule
STAEformer (Original)    & 13.46 & 23.25 & 8.88        \\
STAEformer (Replicated)  & 13.49 & 23.30 & 8.84        \\
+with ST-Denoise (DDPM)          & 13.44 & 23.20 & 8.77     \\
+with ST-LLM (No pretrain)  & 13.45 & \textcolor{red}{\textbf{23.03}} & 8.86    \\
+with ST-LLM (With pretrain) & 14.53    & 24.56     & 10.51  \\
STLLM-DF                   & \textcolor{red}{\textbf{13.38}} & 23.08 &\textcolor{red}{\textbf{ 8.81 }}\\ \bottomrule
\end{tabular}
\label{tab:ablation}
\end{table}

\paragraph{Ablation Study} We conducted a comprehensive ablation study to evaluate the effectiveness of the ST-Denoising block and the ST-LLM block within our proposed STLLM-DF framework (Table~\ref{tab:ablation}). This analysis aims to isolate and assess the contributions of each component, starting from the baseline STAEformer model and progressively incorporating diffusion mechanisms and LLMs, both with and without pertaining.

\begin{figure}[htbp!]
    \centering
    \includegraphics[scale = 0.25]{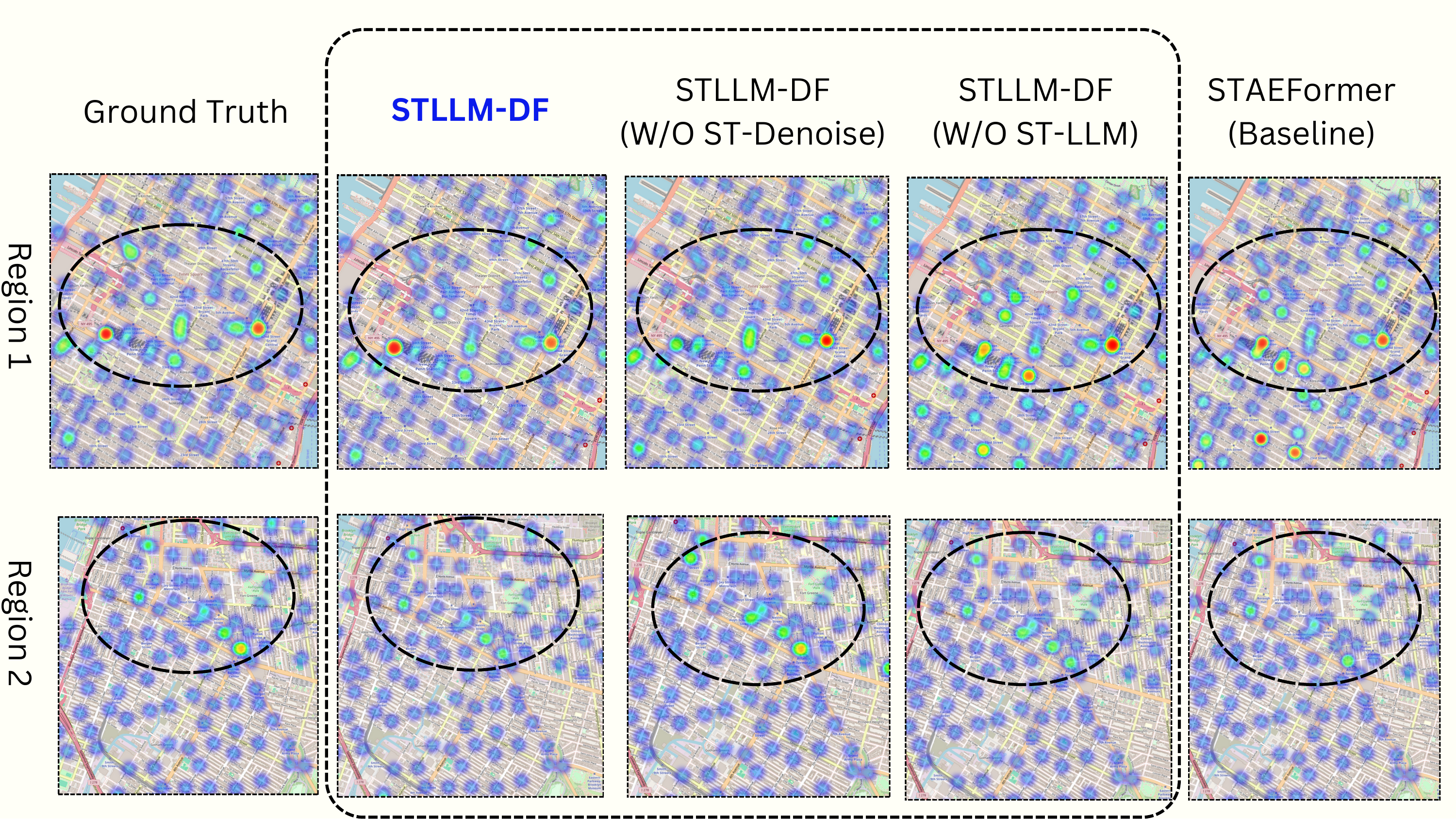}
    \caption{Visualization of traffic predictions for two regions in New York City Bike demand.}
    \label{fig:comparemaps}
\end{figure}

\begin{itemize}
    \item \textbf{ST-Denoising block:} The integration of the diffusion mechanism yields modest but consistent improvements across all performance metrics (MAE: 13.44, RMSE: 23.20, MAPE: 8.77). This enhancement demonstrates the ST-Denoising block's ability to refine predictions by effectively removing noise from the data. The efficacy of DDPM in this context can be attributed to its generative properties: the forward process introduces controlled noise, while the reverse process learns to model the clean data distribution by gradually removing this noise.
    \item \textbf{ST-LLM block - Impact of Pretraining:} The introduction of the ST-LLM block without pretraining shows promising results (MAE: 13.45, RMSE: 23.03, MAPE: 8.86), indicating its capacity to capture complex spatial-temporal dependencies. Surprisingly, the pretrained LLM variant exhibits a performance decline (MAE: 14.53, RMSE: 24.56, MAPE: 10.51). This unexpected result suggests that pretraining may introduce biases incompatible with the specific characteristics of the PEMS08 dataset, leading to reduced generalization.
    \item \textbf{Synergy of ST-Denoising and ST-LLM blocks:} Figure \ref{fig:comparemaps} provides a visual comparison of traffic predictions across different model variants for two distinct regions in New York City. The heatmaps illustrate traffic intensity, with warmer colors indicating higher volumes. Notably, the full STLLM-DF model captures nuanced spatial patterns and hotspots more accurately than its ablated counterparts, particularly in regions with complex traffic dynamics (e.g., Region 1). This visual evidence underscores the complementary roles of the ST-Denoise and ST-LLM components in enhancing prediction accuracy.

\end{itemize}

Our proposed STLLM-DF model, which integrates both the ST-Denoising and ST-LLM blocks, achieves the best overall performance (MAE: 13.38, RMSE: 23.08, MAPE: 8.81). These results demonstrate the model's superior ability to recover the data and learn complex spatial-temporal patterns, leading to more accurate traffic predictions. The diffusion mechanism proves crucial in refining the data, while the LLM block excels in capturing long-range dependencies and facilitating multi-task learning across temporal and spatial domains.

This comprehensive ablation study not only validates the effectiveness of each component but also highlights the synergistic effect of combining ST-Denoising and ST-LLM blocks, resulting in a robust and accurate traffic prediction framework.

\begin{figure}[htbp!]
    \centering
    \includegraphics[scale = 0.5]{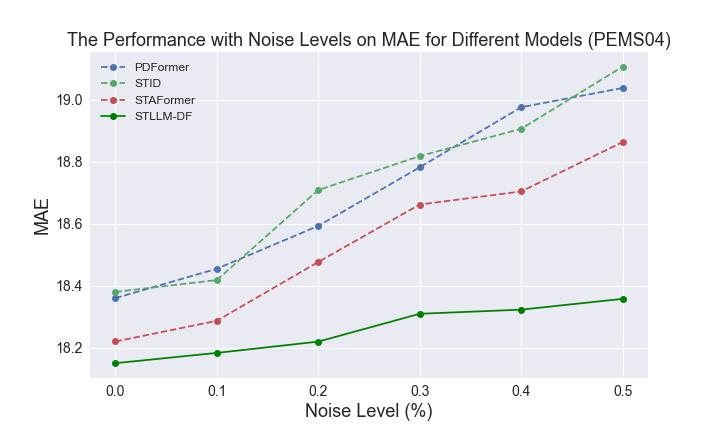}
    \caption{Missing value recovery power analysis for STLLM-DF.}
    \label{fig:noise}
\end{figure}

\paragraph{Denoising Power Analysis} To further analyze the data recovery ability of STLLM-DF, In Figure~\ref{fig:noise} The chart illustrates the performance of four models—PDFormer, STID, STAFormer, and STLLM-DF—on the PEMS04 dataset (with a missing ratio of 3.182\%) across varying missing ratio levels (0.0\% to 0.5\%). The y-axis represents the Mean Absolute Error (MAE), with STLLM-DF demonstrating superior data recovery capability, as indicated by the lowest MAE values across all missing ratio levels. At 0.0\% missing ratio (the clean data), all models perform relatively close, with STLLM-DF still leading slightly. As the missing ratio increases to 0.1\%, 0.2\%, and beyond, the performance gap widens, especially at the higher missing ratio levels of 0.4\% and 0.5\%, where STLLM-DF continues to maintain a low MAE compared to the other models. This observation underscores the effectiveness of STLLM-DF in scenarios where handling missing values is crucial for maintaining prediction accuracy. 

The analysis indicates that STLLM-DF offers superior data recovery capability compared to PDFormer, STID, and STAFormer. Its ability to maintain low MAE across all missing ratio levels makes it an ideal model for applications where data corruption and The analysis indicates that STLLM-DF offers superior data recovery capability compared to PDFormer, STID, and STAFormer. Its ability to maintain low MAE across all missing ratio levels makes it an ideal model for applications where data corruption and missing ratio are prevalent, such as in intelligent transportation systems.  are prevalent, such as in intelligent transportation systems. 

\subsection{Real Case Analysis}
\label{Real Case Analysis}
\begin{figure}[htbp!]
    \centering
    \includegraphics[scale = 0.3]{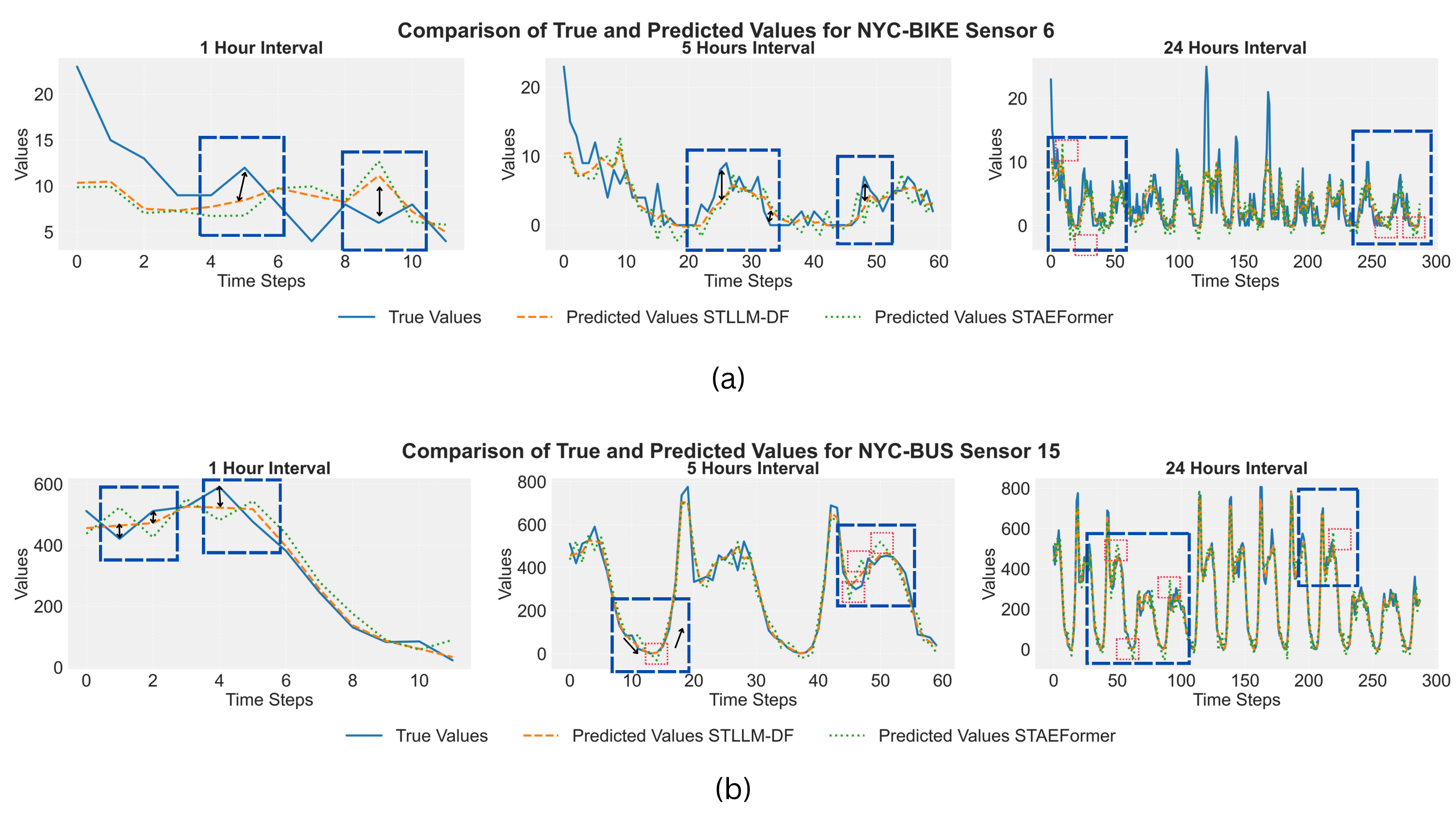}
    \caption{Comparative Analysis of Prediction Results Using NYC-BIKE and NYC-BUS Dataset for Sensors 6 and
15.}
    \label{fig:bikebus}
\end{figure}

In this section, we evaluate the predictive power of STLLM-DF competing with the formal SOTA model STAEFormer. In Figure~\ref{fig:bikebus}, a comparison of True and Predicted Values for NYC-BIKE Sensor 6 and NYC-BUS Sensor 15 using STLLM-DF and STAFormer models. Subfigure (a) illustrates the performance of the models for NYC-BIKE Sensor 6 across 1-hour, 5-hour, and 24-hour intervals. Subfigure (b) shows a similar comparison for NYC-BUS Sensor 15 across the same intervals. The blue solid line represents the true values, while the dashed orange line and dotted green line represent the predicted values from STLLM-DF and STAFormer, respectively. The blue boxes highlight specific regions where STLLM-DF more accurately follows the true values, particularly during periods of sharp fluctuations and peaks, demonstrating its superior predictive and data recovery capabilities.

\begin{figure}[htbp!]
    \centering
    \includegraphics[scale = 0.3]{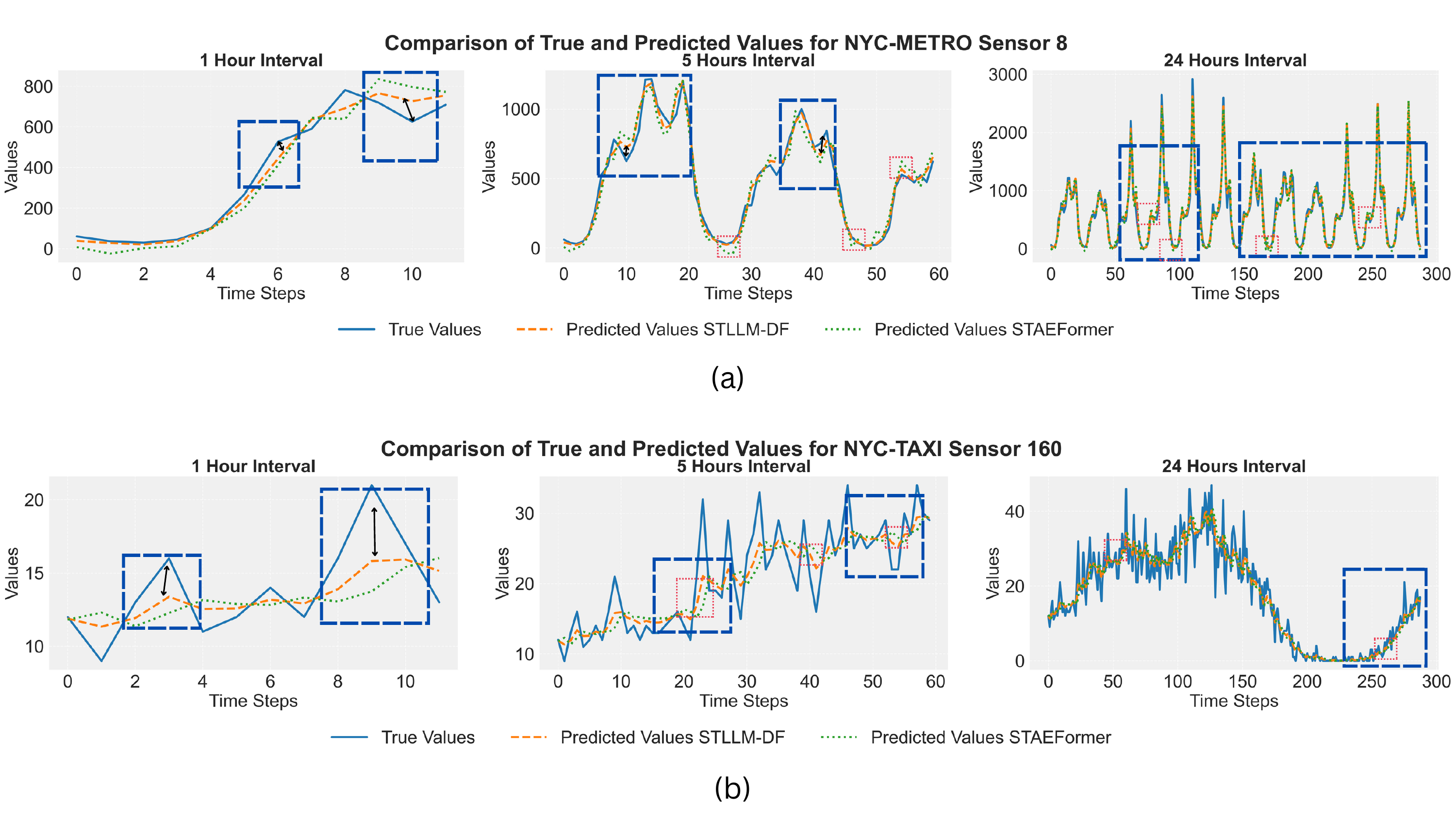}
    \caption{Comparative Analysis of Prediction Results Using NYC-METRO and NYC-TAXI Dataset for Sensors 8 and
160.}
    \label{fig:METAXI}
\end{figure}

In Figure~\ref{fig:bikebus}, the comparative analysis of prediction results for NYC-BIKE and NYC-BUS datasets reveals the superior performance of STLLM-DF over STAEFormer across various time intervals. For NYC-BIKE (Sensor 6), STLLM-DF demonstrates more accurate predictions, especially during peak hours. In the 24-hour interval, STLLM-DF captures daily patterns more precisely, predicting approximately 7 bikes at step 250, closer to the true value of 8, while STAEFormer underestimates at around 5. The NYC-BUS (Sensor 15) results further highlight STLLM-DF's strengths, particularly in predicting high-volume periods. At step 275 in the 24-hour interval, STLLM-DF accurately predicts 700 passengers, matching the true value, whereas STAEFormer underestimates at about 600. Both models perform well in capturing general trends, such as the sharp decline from 600 to 100 passengers over 10 steps in the 1-hour interval for NYC-BUS. However, STLLM-DF consistently outperforms in capturing both short-term fluctuations and long-term patterns, which is especially evident in the 5-hour and 24-hour intervals across both datasets. This analysis underscores STLLM-DF's enhanced capability to handle complex urban transportation data with varying temporal granularities. The performance of the STLLM-DF model is evaluated against the STAFormer model across multiple sensors in Figure~\ref{fig:METAXI}; the comparisons focus on true and predicted values for both NYC-METRO Sensor 8 and NYC-TAXI Sensor 160 across different time intervals: 1-hour, 5-hour, and 24-hour intervals.

Figure \ref{fig:METAXI}(a) highlights the model performance for NYC-METRO Sensor 8. Across the three time intervals, STLLM-DF consistently aligns closely with the true values, particularly in capturing the peaks and troughs in the 5-hour and 24-hour intervals. In comparison, the STAFormer model exhibits similar patterns but struggles slightly in accurately predicting rapid fluctuations, as seen in the 5-hour interval plot. Notably, in the 24-hour interval, both models perform well, but STLLM-DF appears to better follow the intricate spikes and drops in the data, especially in the highlighted sections where sudden increases and decreases occur. This suggests that STLLM-DF excels in handling highly dynamic and complex traffic patterns typical in urban metro systems.

Figure~\ref{fig:METAXI}(b) presents the results for NYC-TAXI Sensor 160, where the models are again evaluated over 1-hour, 5-hour, and 24-hour intervals. Similar to the metro data, STLLM-DF outperforms STAFormer in closely tracking the true values, particularly during high variation periods. For instance, in the 5-hour interval, STLLM-DF is better at following the rising trend around time step 50, as shown in the highlighted area. STAFormer slightly underestimates this trend, indicating a lag in capturing the rapid increase. In the 24-hour interval, STLLM-DF maintains a tighter correlation with the true values, especially in areas where the data shows steep declines followed by gradual rises.  The STAFormer model, while performing admirably, shows more pronounced deviations in these complex regions, indicating that it may struggle with handling highly fluctuating traffic volumes.

\section{Discussion and Implementation}

\label{Discussion and Implementation}
\paragraph{Discussion} Our study demonstrates that the STLLM-DF model consistently outperforms existing state-of-the-art models across various datasets, such as PEMS and NYC transportation, when evaluated using metrics like MAE, RMSE, and MAPE. The model’s strong performance, especially in complex scenarios and on both short and long prediction periods, shows that it can generalize well to different settings. Moreover, its effectiveness across multiple transportation modes, including bikes, buses, taxis, and metros, highlights its versatility in handling diverse urban mobility data.

\begin{figure}[htbp!]
    \centering
    \includegraphics[scale = 0.2]{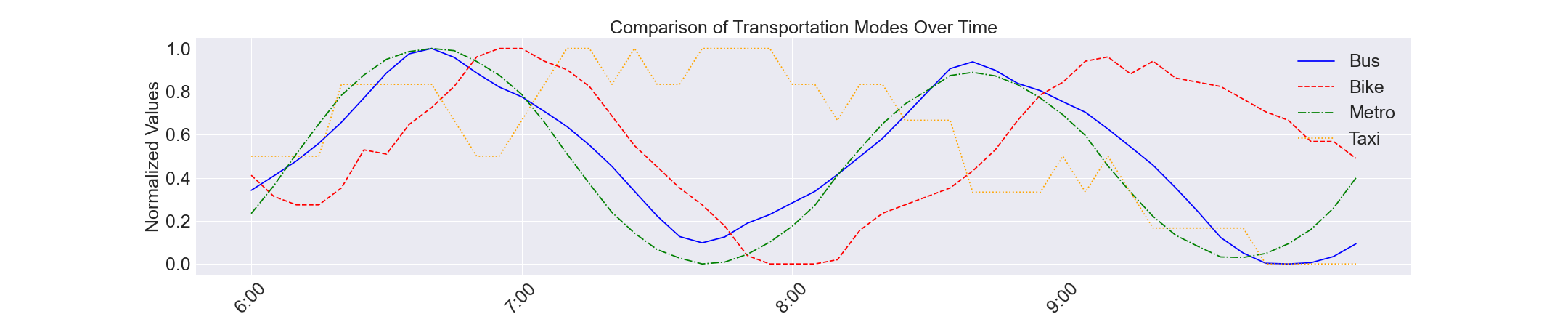}
    \caption{Comparison of public transportation modes overtime for New York City.}
    \label{fig:comparelines}
\end{figure}

\paragraph{Implementation and Insight}The demo prediction results, as illustrated in Figure~\ref{fig:comparelines}, provide a clear comparison of different transportation modes (bus, bike, metro, and taxi) during a typical weekday. The data reveal a peak in demand, particularly around 7:00 AM and 9:00 AM, underscoring the importance of efficient resource management during rush hours. Buses and metros show similar peak usage patterns, indicating their central role in public transportation. In contrast, bikes and taxis demonstrate more varied demand, reflecting their more flexible and individualized use.

These intermodal dynamics underscore the necessity of a centralized Intelligent Transportation System (ITS), as depicted in Figure~\ref{fig:intro}. The complementary relationship between modes, such as the inverse correlation between bus and bike usage around peak times, highlights opportunities for a centralized ITS to optimize the transportation network. For instance, as bus usage decreases post-rush hour, a system could promote bike use by ensuring adequate availability of bikes or dedicated lanes. Conversely, during adverse weather, the ITS could dynamically allocate more buses to absorb the shift away from cycling.

Our STLLM-DF model’s ability to simultaneously analyze data from multiple transportation modes allows for these critical intermodal insights, supporting system-wide optimization efforts. By handling tasks such as demand forecasting, traffic flow prediction, and arrival time estimation, our model enhances urban transportation management. It provides essential tools for optimizing resource allocation, reducing congestion, and mitigating delays during peak periods, contributing to the overall efficiency of centralized ITS.

\section{Conclusions}
\label{conclusion}
This paper focuses on enhancing multi-task transportation prediction within Intelligent Transportation Systems (ITS) through a novel Spatial-Temporal Large Language Model Diffusion (STLLM-DF) model. The main conclusions drawn from this work are summarized as follows:
\begin{itemize}
    \item \textbf{Performance Improvement through Multi-Task Learning:} The comparisons between STLLM-DF and traditional single-task models highlight the advantage of multi-task learning in ITS. Jointly considering diverse transportation modes allows for better prediction performance across multiple systems, including buses, taxis, and metros, by leveraging shared spatial-temporal information.
    \item \textbf{Innovative Model Components: }The integration of Denoising Diffusion Probabilistic Models (DDPMs) and Large Language Models (LLMs) in the STLLM-DF framework is capable of adaptively extracting and refining noisy data. These components effectively enhance the spatial and temporal interactions among multi-traffic modes, improving prediction accuracy and robustness.
    \item \textbf{Extensive Validation:} The superiority of STLLM-DF is confirmed through extensive experiments. In comparison to baseline models, STLLM-DF demonstrated consistent improvements, with a 2.40\% reduction in MAE, 4.50\% reduction in RMSE, and 1.51\% reduction in MAPE, showcasing its strong performance in handling real-world noisy transportation data.
\end{itemize}

Nevertheless, this paper has some limitations. First, while STLLM-DF focuses on multi-traffic modes, the current scope is limited to a few modes (buses, taxis, metros, bike-sharing). In reality, ITS includes more diverse transportation systems, such as carpooling, and more, which could further evaluate the model’s performance. Second, external data sources such as weather conditions, social events, and infrastructure variations, which often impact transportation systems, have not been incorporated into the model due to data acquisition challenges.

Future work will address these limitations by expanding the model to include additional traffic modes and external data sources, thus creating a more comprehensive multi-modal transportation system. Additionally, we will apply STLLM-DF to various challenging scenarios, such as emergency responses and extreme weather conditions. Finally, future research will focus on improving the interpretability of the model, particularly by quantifying the interdependencies between different transportation modes.
\newpage

\appendix




\bibliographystyle{elsarticle-harv}
\bibliography{example}





\end{document}